\pdfoutput=1

\documentclass[11pt]{article}

\usepackage{ACL2023}

\usepackage{times}
\usepackage{latexsym}

\usepackage[T1]{fontenc}

\usepackage[utf8]{inputenc}

\usepackage{microtype}

\usepackage{inconsolata}

\usepackage{graphicx}
\usepackage{amsmath}
\usepackage{adjustbox}
\usepackage{booktabs}
\usepackage{multirow}
\usepackage{tcolorbox}
\usepackage{lipsum}
\usepackage{todonotes}
\usepackage{listings}
\usepackage{gensymb}

\title{The Illusion of Progress: Re-evaluating Hallucination Detection in LLMs}

\author{
\textbf{Denis Janiak}\textsuperscript{1} \quad
\textbf{Jakub Binkowski}\textsuperscript{1} \quad
\textbf{Albert Sawczyn}\textsuperscript{1} \quad
\textbf{Bogdan Gabrys}\textsuperscript{2} \\
\textbf{Ravid Shwartz-Ziv}\textsuperscript{3} \quad
\textbf{Tomasz Kajdanowicz}\textsuperscript{1} \\
\textsuperscript{1}Wroclaw University of Science and Technology \\
\textsuperscript{2}University of Technology Sydney \\
\textsuperscript{3}New York University
}

\begin{document}
\maketitle

\begin{abstract}
Large language models (LLMs) have revolutionized natural language processing, yet their tendency to hallucinate poses serious challenges for reliable deployment. Despite numerous hallucination detection methods, their evaluations often rely on ROUGE, a metric based on lexical overlap that misaligns with human judgments. Through comprehensive human studies, we demonstrate that while ROUGE exhibits high recall, its extremely low precision leads to misleading performance estimates. In fact, several established detection methods show performance drops of up to 45.9\% when assessed using human-aligned metrics like LLM-as-Judge. Moreover, our analysis reveals that simple heuristics based on response length can rival complex detection techniques, exposing a fundamental flaw in current evaluation practices. We argue that adopting semantically aware and robust evaluation frameworks is essential to accurately gauge the true performance of hallucination detection methods, ultimately ensuring the trustworthiness of LLM outputs.
\end{abstract}

\section{Introduction}

\begin{figure}[h!]
    \centering
    \includegraphics[width=0.49\textwidth]{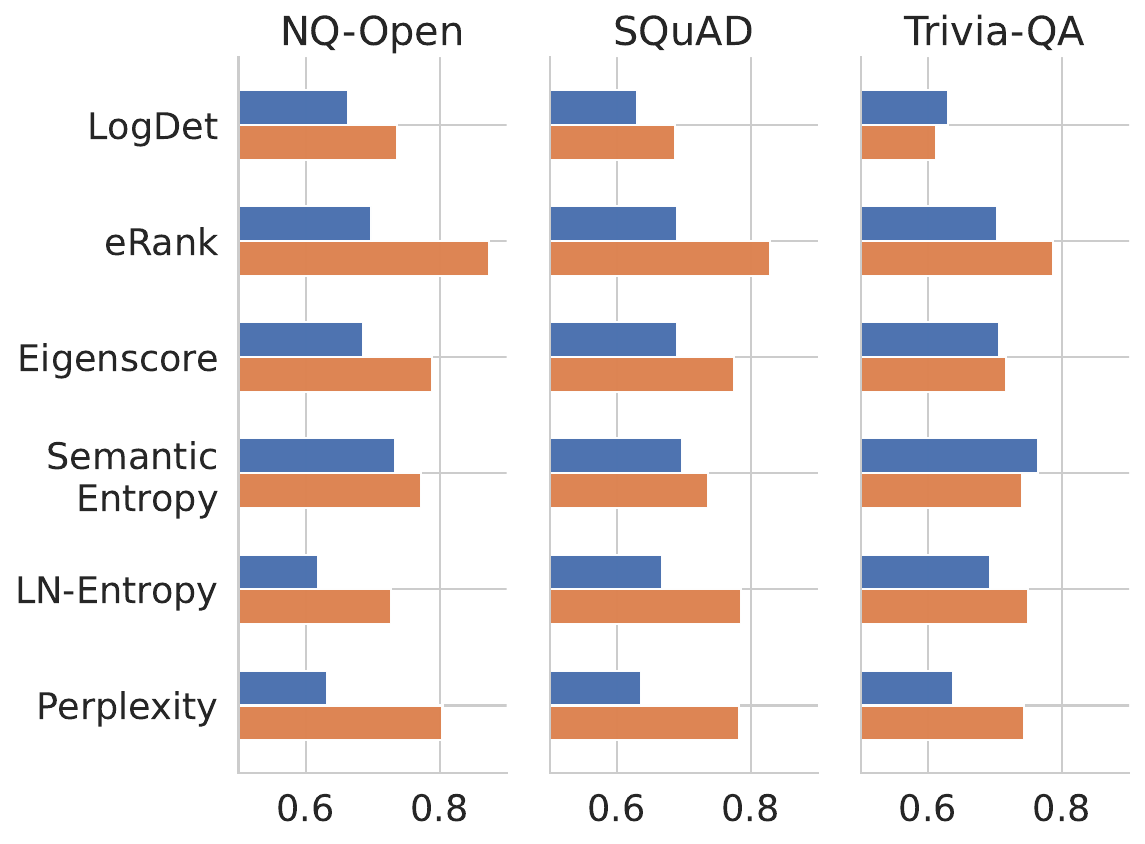}
    \caption{\textbf{ROUGE-based evaluation fails to reliably capture true hallucination detection capabilities.} Hallucination detection performance (AUROC) comparison of \textcolor{orange}{\textbf{ROUGE-L}} and \textcolor{blue}{\textbf{LLM-as-Judge}} evaluation across three datasets. Many methods shows significant evaluation discrepancies.}
    \label{fig:rouge_vs_llm_mistral_few_shot}
\end{figure}

Large language models (LLMs) have transformed natural language processing, but their tendency to hallucinate—generating fluent yet factually incorrect outputs—poses a \textbf{critical challenge} for real-world applications \cite{lei_2025_survey}.
As LLMs are increasingly deployed in high-stakes scenarios, unsupervised hallucination detection has emerged as a promising solution, offering scalable evaluation without the generalization limitations of supervised approach and costly annotation process \cite{su_unsupervised_2024}.
A growing body of work has explored this direction \citep{chen_inside_2024, farquhar_detecting_2024, du_haloscope_2024, kernel_language_entropy, semantic_density, duan-etal-2024-shifting, nguyen2025smoothinghallucinationsmitigatingllm}, often relying on ROUGE as the primary correctness metric.
ROUGE, originally developed to assess summary quality based on lexical overlap \citep{lin_rouge_2004}, is used to approximate factual consistency by applying threshold-based heuristics: responses with low ROUGE overlap to reference answers are often labeled as hallucinated.
However, the suitability of ROUGE for assessing the factual accuracy of Question Answering (QA) responses, specifically in identifying hallucinations, has been
largely assumed rather than rigorously validated.

Existing critiques of ROUGE often focus on its limitations in capturing fluency or adequacy in long-form summarization or dialogues \cite{honovich_true_2022, dziri-etal-2022-faithdial, zhong-etal-2022-unieval}. In contrast, this paper presents a \textbf{systematic, large-scale empirical investigation} specifically evaluating ROUGE's efficacy in the context of QA hallucination detection.
Our analysis goes beyond general critiques by quantitatively demonstrating ROUGE’s key shortcomings—such as its susceptibility to response length—and how these issues can inflate the reported performance of hallucination detection methods.
Furthermore, while ROUGE serves as our primary case study due to its ubiquity, we also demonstrate that other commonly used metrics, including those based on n-grams and semantic similarity, share similar vulnerabilities in this specific task, highlighting a broader deficiency in current evaluation practices.

To establish a human‑aligned benchmark, we collect human judgments of factual correctness and compare metric outputs against these gold labels. We find that ROUGE exhibits \textbf{alarmingly low precision} for identifying actual factual errors. In contrast, an LLM‑as‑Judge approach \cite{zheng_judging_2023} aligns far more closely with human assessments. Based on these insights, we re‑evaluate existing detection methods under both ROUGE and human‑aligned criteria, revealing dramatic performance drops (up to 45.9\% for Perplexity and 30.4\% for Eigenscore) when moving from ROUGE to LLM‑as‑Judge evaluation (see Figure~\ref{fig:rouge_vs_llm_mistral_few_shot}).

Finally, we uncover a surprising baseline: simple length‑based heuristics (e.g., mean and standard deviation of answer length) rival or exceed sophisticated detectors like Semantic Entropy. Through controlled experiments that isolate length effects, we show how ROUGE can be manipulated via trivial repetition, even when factual content remains constant. Our findings expose a \textbf{widespread overestimation} of current methods and underscore the urgent need for more reliable, human‑aligned evaluation metrics in QA hallucination detection.

Our study makes the following key contributions:
\begin{enumerate}
    \item A human evaluation study validating LLM‑as‑Judge as a reliable metric for factual correctness, while demonstrating that ROUGE—and other n‑gram and semantic metrics—are severely misaligned with human judgments.
    \item A systematic re-evaluation of existing hallucination detection methods, showing that their effectiveness is often overstated when assessed with ROUGE and similar metrics, and revealing how these metrics can hide important flaws in the methods.
    \item Evidence that response length is a surprisingly effective indicator of hallucination, with simple length-based heuristics often matching or exceeding the performance of more sophisticated detection approaches.
\end{enumerate}

\section{Related Work}
\paragraph{Hallucination Detection Methods}
Recent research has shown that hallucinations in LLMs are inevitable \citep{xu_hallucination_2024}, spurring work on two main detection paradigms: supervised and unsupervised.
\emph{Supervised} methods usually employ probing classifiers trained on labeled hidden states to detect hallucinations \citep{azaria_internal_2023,orgad_llms_2024,arteaga_hallucination_2024}. While effective, they depend on costly human annotations and often fail to generalize across domains.
\emph{Unsupervised} methods detect hallucinations by estimating uncertainty directly—token‑level confidence from single generations \citep{ren2023outofdistribution}, sequence‑level variance across multiple samples \citep{malinin2021uncertainty,farquhar_detecting_2024}, or hidden‑state pattern analysis \citep{chen_inside_2024,llm_check}.
While these methods show strong performance on standard benchmarks, our analysis reveals that simpler length‑based baselines can achieve comparable results—echoing prior findings that simple baselines remain surprisingly competitive and underscoring the need for rigorous head‑to‑head comparisons \citep{fadeeva-etal-2023-lm}.

\paragraph{Evaluation Metrics and Their Limitations} Traditional n‑gram overlap measures such as ROUGE \citep{lin_rouge_2004} remain popular for detecting hallucinations, despite their inability to reliably assess factual consistency \citep{honovich_true_2022}. Recent studies have further highlighted these limitations, particularly in multilingual settings where lexical overlap proves unreliable compared to NLI-based approaches \citep{kang2024}. Even ROUGE‑L, which tracks the longest common subsequence, often misses errors that leave surface overlap intact. To overcome these shortcomings, a family of embedding‑based metrics — BERTScore \cite{bertscore}, UniEval \cite{zhong-etal-2022-unieval}, AlignScore \citep{zha-etal-2023-alignscore}, and related approaches — has been proposed to capture deeper semantic similarity. However, these learned representations can still diverge from human judgments of truthfulness. By contrast, LLM-as-Judge methods \citep{zheng_judging_2023} have shown strong agreement with human judgments in QA tasks \citep{thakur2025judgingjudgesevaluatingalignment}, offering a more reliable alternative. Our study builds on these insights by exposing ROUGE’s and other metrics blind spots and validating LLM‑as‑Judge as a more faithful framework for factual evaluation.

\section{Experimental Setup}
\subsection{Overview}
Our experimental design aims to investigate both the shortcomings of current evaluation methods and the effectiveness of simpler alternatives.

\subsection{Datasets and Models}
For our experiments, we use three established QA datasets, each with distinct characteristics:

\begin{itemize}
    \item \textbf{NQ-Open} \citep{kwiatkowski_natural_2019}: Contains 3,610 question-answer pairs drawn from real Google search queries, representing natural information-seeking behavior
    \item \textbf{TriviaQA} \citep{joshi_triviaqa_2017}: A subset of 3,842 examples from the validation set, featuring trivia questions that often require specific factual knowledge
    \item \textbf{SQuAD} \citep{rajpurkar_know_2018}: 4,150 examples from the validation set (rc.nocontext), characterized by longer, more complex questions and answers
\end{itemize}

NQ-Open and TriviaQA primarily feature shorter questions and answers, whereas SQuADv2 contains longer inputs, making it suitable for evaluating our method in more complex contexts.

We generated answers using two open-source LLMs: \textsc{Llama3.1-8B-Instruct}\footnote{\href{https://hf.co/meta-llama/Llama-3.1-8B-Instruct}{hf.co/meta-llama/Llama-3.1-8B-Instruct}} \citep{grattafiori_llama_2024} and \textsc{Mistral-7B-Instruct-v0.3}\footnote{\href{https://huggingface.co/mistralai/Mistral-7B-Instruct-v0.3}{hf.co/mistralai/Mistral-7B-Instruct-v0.3}} \citep{jiang_mistral_2023}. For simplicity, we refer to these models as \textsc{Llama} and \textsc{Mistral} in our plots and tables.

\subsection{Hallucination Detection Baselines}
We compare our approach against established baselines that fall into two categories. \textbf{Uncertainty-based methods} estimate model confidence, including \texttt{Perplexity} \citep{ren2023outofdistribution}, Length-Normalized Entropy (\texttt{LN-Entropy}) \citep{malinin2021uncertainty}, and Semantic Entropy (\texttt{SemEntropy}) \citep{farquhar_detecting_2024}, which use multiple generations to capture sequence-level uncertainty.
\textbf{Consistency-based methods} analyze internal representations. \texttt{EigenScore} \citep{chen_inside_2024} computes generation consistency via eigenvalue spectra, while \texttt{LogDet} \citep{llm_check} measures covariance structure from single generations.
We also evaluate Effective Rank (\texttt{eRank}) \citep{roy_effective_2007, garrido_rankme_2023}, an intrinsic dimensionality measure we adapt as a novel hallucination indicator (see Appendix \ref{app:erank}).

\subsection{Ground Truth Labels}
To obtain reliable ground truth labels for evaluating the correctness of generated responses, we utilize two complementary approaches:

\textbf{LLM-as-Judge}
leverages GPT-4o-Mini \citep{openai_gpt-4_2024} for semantic assessment, following the methodology outlined in \citep{zheng2023judging} and using a prompt adapted from \citep{orgad2025llms}. This approach classifies generated responses into three categories: "correct," "incorrect," or "refuse" (with "refuse" being treated as a hallucination). By focusing on semantic equivalence and factual accuracy, this method goes beyond surface-level comparisons and exhibits strong alignment with human judgments \cite{thakur2025judgingjudgesevaluatingalignment}.

\textbf{ROUGE-L F1 Score}
\citep{lin_rouge_2004} measures the longest common subsequence between the generated response and the ground truth. Consistent with prior work \citep{farquhar_detecting_2024}, we apply a threshold of 0.3 for this metric. Including ROUGE-L allows us to compare our findings with existing literature and highlight the limitations of relying solely on lexical overlap for evaluating factual correctness. It helps to quantify the discrepancy between semantic understanding (assessed by the LLM judge) and simple word matching.

\subsection{Evaluation Metrics}
We employ Area Under the Receiver Operating Characteristic curve (AUROC) and Area Under the Precision-Recall curve (PR-AUC) as our primary evaluation metrics. AUROC assesses the ability of a hallucination detection method to correctly rank positive and negative instances (hallucinations vs. non-hallucinations). PR-AUC is particularly valuable when dealing with imbalanced datasets, which is often the case in hallucination detection where non-hallucinated responses might be more frequent. Both metrics offer a threshold-independent evaluation of the ranking performance \citep{lin_generating_2023}.

\subsection{Implementation Details}
We utilize pretrained model weights from the Hugging Face Transformers \citep{wolf_transformers_2020} without any additional fine-tuning. Following \citep{farquhar_detecting_2024}, we generate 10 samples ($n=10$) using temperature $1.0$ for uncertainty estimation. Additionally, we generate one "best answer" sample with temperature  $0.1$ to serve as the best-generation estimate for performance evaluation.

The models are evaluated in both zero-shot and few-shot ($k=5$) settings:
\begin{itemize}
    \item \textbf{Zero-shot}: Models rely solely on their pre-existing knowledge, testing base capabilities
    \item \textbf{Few-shot}: Models receive five carefully selected examples demonstrating expected answer formats
\end{itemize}

Both settings use a standardized prompt designed to elicit concise answers. The specific prompt, adapted from \citep{kossen_semantic_2024}, can be found in Appendix \ref{app:prompts}. We report results for a single run unless specified otherwise.

\section{Human Evaluation: The Gold Standard}
\label{sec:human_eval}
Before analyzing  the technical problems of hallucination detection methods, we first establish that commonly used evaluation metrics—specifically ROUGE—are poorly aligned with human judgments of factual correctness \cite{honovich_true_2022, kang2024}. In contrast, an evaluation method based on LLM-as-Judge demonstrates much closer agreement with human assessments \cite{thakur2025judgingjudgesevaluatingalignment}. To illustrate this, we conducted a comprehensive human evaluation study.

\paragraph{Study Design}
We randomly selected 200 question--answer pairs from the Mistral answers on the NQ-Open dataset, ensuring a balanced representation of cases where ROUGE and LLM-as-Judge yield conflicting hallucination assessments. Each answer was independently assessed by three annotators using standardized guidelines from \cite{thakur2025judgingjudgesevaluatingalignment}, classifying responses as \emph{correct}, \emph{incorrect}, or \emph{refuse} (we then classify model refusal as incorrect). The high inter-annotator agreement (Cohen's Kappa = 0.799) confirms the reliability of human judgments.

\paragraph{Key Findings}
Our results reveal a significant performance gap between LLM-as-Judge and ROUGE when benchmarked against human consensus. While ROUGE exhibits high precision but fails to detect many hallucinations, LLM-as-Judge achieves significantly higher recall, aligning more closely with human assessments, as shown in Table \ref{tab:human_agreements}.

\begin{table}[h!]
\centering
\caption{\textbf{LLM-as-Judge provides superior alignment with human judgment.} Comparison of ROUGE (with standard 0.3 threshold) and LLM-as-Judge against human labels.}
\label{tab:human_agreements}
\adjustbox{max width=0.5\textwidth}{
\begin{tabular}{lrrrr}
\toprule
Method & Precision & Recall & F1-Score & Agreement \\
\midrule
LLM-as-Judge & 0.736 & 0.957 & 0.832 & 0.723 \\
ROUGE & 0.401 & 0.957 & 0.565 & 0.142 \\
\bottomrule
\end{tabular}
}
\end{table}

\paragraph{Implications}

Our findings underscore that ROUGE is a poor proxy for human judgment in evaluating hallucination detection. Despite its high precision, ROUGE fails to capture many critical errors, resulting in a significant misalignment with human assessments of factual correctness. In contrast, LLM-as-Judge exhibits strong agreement with human evaluations—achieving both high precision and recall—which motivates its adoption as a more robust, semantically aware evaluation method throughout this work.

\section{Re-evaluating Hallucination Detection Methods}
\label{sec:results}

\subsection{Limitations of ROUGE for Factual Accuracy Assessment in QA}
\label{sec:rouge_limitations_summary}

The predominant reliance on ROUGE for evaluating QA hallucination detection methods warrants careful scrutiny, as its core design for lexical overlap does not inherently capture factual correctness. Our in-depth analysis, presented in Appendix \ref{app:rouge_failure_modes_analysis}, reveals several critical failure modes that systematically undermine ROUGE's utility for this task. Key limitations include: sensitivity to response length, inability to handle semantic equivalence and susceptibility to false lexical matches.

\begin{figure}[h!]
    \includegraphics[width=\linewidth]{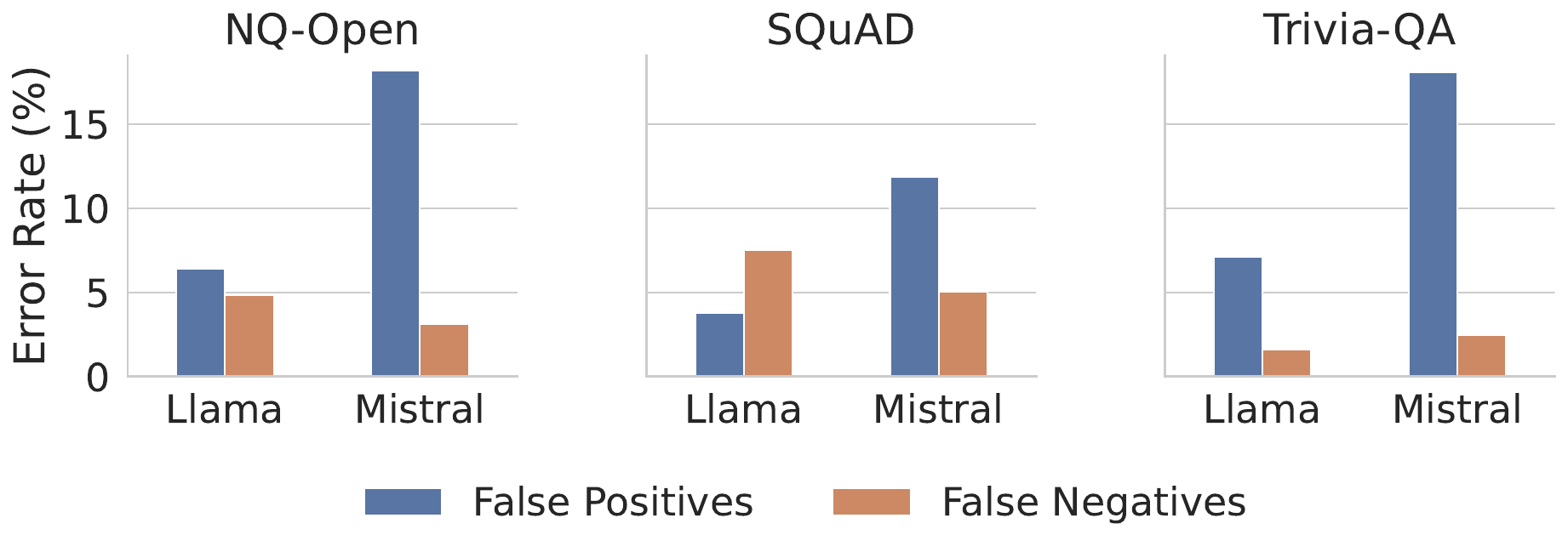}
    \caption{\textbf{ROUGE produces systematic errors across all evaluation settings.} Distribution of \textit{False Negatives} and \textit{False Positives} across different datasets and models highlights the inconsistency in ROUGE's evaluation. }
    \label{fig:fn_fp_barplot}
\end{figure}

\begin{table*}[ht!]
    \centering
        \caption{\textbf{Detection methods show dramatic performance drops when evaluated against human-aligned metrics instead of ROUGE.} Performance comparison using AUROC scores for \textsc{llama} and \textsc{mistral} models across three datasets in zero-shot setting, where negative $\Delta\%$ values reveal ROUGE's overestimation of method effectiveness.}
    \adjustbox{valign=c, max width=\textwidth, scale=1}{\begin{tabular}{ll ccc @{\hspace{2em}} ccc @{\hspace{2em}} ccc @{\hspace{2em}} ccc}
\toprule
\multirow{2}{*}{Model} & \multirow{2}{*}{Metric} & \multicolumn{3}{c}{NQ-Open} & \multicolumn{3}{c}{SQuAD} & \multicolumn{3}{c}{Trivia-QA} & \multicolumn{3}{c}{Mean} \\
                        &                         & ROUGE & LLM & $\Delta$\% & ROUGE & LLM & $\Delta$\%  & ROUGE & LLM & $\Delta$\%  & ROUGE & LLM & $\Delta$\%  \\
\midrule \midrule
\multirow{5}{*}{\textsc{Llama}} & \texttt{Perplexity} & 0.709 & 0.700 & -1.2 & 0.703 & 0.687 & -2.4 & 0.733 & 0.789 & 7.2 & 0.715 & 0.725 & 1.2 \\
 & \texttt{LN-Entropy} & 0.521 & 0.605 & 13.9 & 0.558 & 0.611 & 8.7 & 0.563 & 0.636 & 11.5 & 0.547 & 0.617 & 11.4 \\
 & \texttt{SE} & 0.778 & 0.742 & -4.8 & 0.707 & 0.705 & -0.2 & 0.769 & 0.832 & 7.6 & 0.751 & 0.760 & 0.9 \\
 & \texttt{Eigenscore} & 0.816 & 0.686 & -19.0 & 0.720 & 0.638 & -12.7 & 0.752 & 0.734 & -2.5 & 0.763 & 0.686 & -11.4 \\
 & \texttt{eRank} & 0.825 & 0.632 & -30.6 & 0.754 & 0.621 & -21.4 & 0.717 & 0.660 & -8.6 & 0.765 & 0.638 & -20.2 \\ 
  & \texttt{LogDet} & 0.511 & 0.515 & 0.7 & 0.521 & 0.536 & 2.7 & 0.604 & 0.509 & -18.6 & 0.545 & 0.520 & -5.1 \\ \hline
  \multirow{5}{*}{\textsc{Mistral}} & \texttt{Perplexity} & 0.852 & 0.584 & -45.9 & 0.516 & 0.500 & -3.2 & 0.843 & 0.627 & -34.4 & 0.737 & 0.570 & -27.8 \\
 & \texttt{LN-Entropy} & 0.718 & 0.645 & -11.3 & 0.734 & 0.657 & -11.7 & 0.586 & 0.596 & 1.8 & 0.679 & 0.633 & -7.1 \\
 & \texttt{SE} & 0.836 & 0.729 & -14.7 & 0.784 & 0.701 & -11.9 & 0.726 & 0.707 & -2.6 & 0.782 & 0.712 & -9.7 \\
 & \texttt{Eigenscore} & 0.873 & 0.669 & -30.4 & 0.803 & 0.648 & -24.0 & 0.775 & 0.652 & -18.9 & 0.817 & 0.656 & -24.4 \\
 & \texttt{eRank} & 0.925 & 0.678 & -36.4 & 0.518 & 0.511 & -1.3 & 0.851 & 0.645 & -31.9 & 0.765 & 0.611 & -23.2 \\
 & \texttt{LogDet} & 0.628 & 0.508 & -23.6 & 0.562 & 0.518 & -8.5 & 0.843 & 0.606 & -39.2 & 0.678 & 0.544 & -23.8 \\
\bottomrule
\end{tabular}
}

    \label{tab:metrics_results_zero_shot}
\end{table*}

These failure modes, illustrated with concrete examples and error distributions in Figure \ref{fig:fn_fp_barplot}, highlight the potential for ROUGE to provide a misleading assessment of both LLM responses and the efficacy of hallucination detection techniques. This underscores the need for evaluation against more human-aligned metrics.

\subsection{Quantifying the Evaluation Gap: ROUGE vs. LLM-as-Judge}
\label{sec:quantifying_evaluation_gap}

Given the outlined limitations of ROUGE, we re-evaluated existing unsupervised hallucination detection methods using LLM-as-Judge, which, as validated by our human study, offers a closer alignment with human judgments of factual correctness.

\paragraph{Main results} As detailed in Table~\ref{tab:metrics_results_zero_shot}, hallucination detection methods that show promise under ROUGE often suffer a substantial performance drop when re-evaluated with LLM-as-Judge. For instance, \texttt{Perplexity} sees its AUROC score plummet by as much as \textbf{45.9\%} for the \textsc{Mistral} model on NQ-Open. Similarly, \texttt{Eigenscore}'s performance erodes by \textbf{19.0\%} and \textbf{30.4\%} for \textsc{Llama} and \textsc{Mistral}, respectively, on the same dataset. Even \texttt{eRank}, which posts impressive ROUGE-based scores, experiences a sharp decline of \textbf{30.6\%} and \textbf{36.4\%} under the LLM-as-Judge paradigm. Moreover, when evaluated using PR-AUC, we observe even larger performance discrepancies across all methods (see Tables~\ref{tab:appendix_prauc_metrics_results_zero_shot_full} and~\ref{tab:appendix_prauc_metrics_results_few_shot_full} in the Appendix \ref{app:prauc}); this amplifies the impact of class imbalance in the QA setup, as further evidenced by the low QA accuracies reported in Table~\ref{tab:appendix_accuracy}.

\begin{figure}[h!]
    \includegraphics[width=\linewidth]{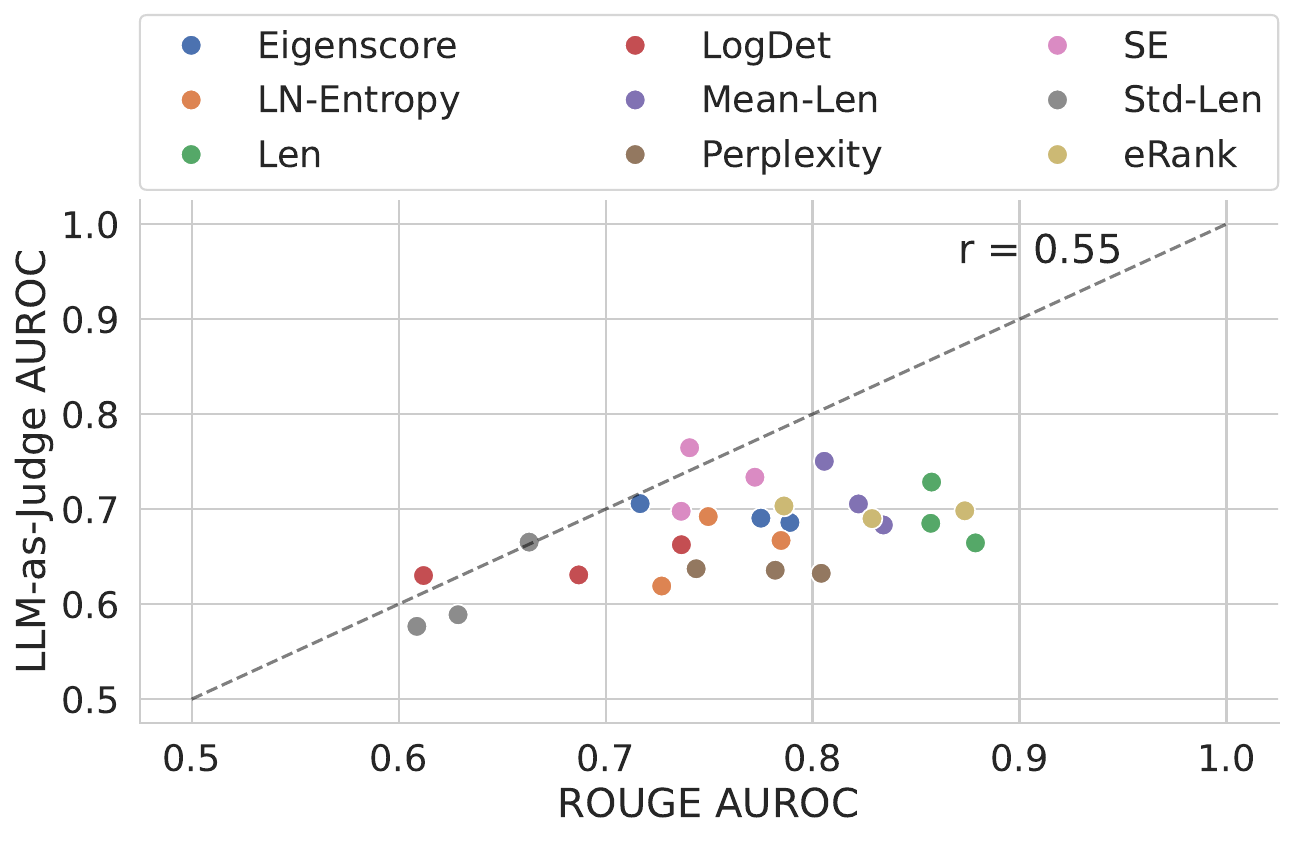}
    \caption{\textbf{ROUGE and human-aligned evaluations show weak correlation across detection methods.} Correlation between ROUGE and LLM-as-Judge AUROC scores for the \textsc{mistral} model, with each point representing a metric's performance on specific dataset.}

    \label{fig:rouge_vs_llm_corr_mistral_few_shot}
\end{figure}

\paragraph{Correlation} This systematic discrepancy, visually underscored by the scatter plot in Figure~\ref{fig:rouge_vs_llm_corr_mistral_few_shot}, points to a fundamental inadequacy in ROUGE's ability to reflect true hallucination detection performance. The moderate Pearson correlation coefficient ($r = 0.55$) between the AUROC scores derived from these two evaluation approaches further suggests that methods may be inadvertently optimized for ROUGE's lexical overlap criteria rather than genuine factual correctness. Notably, among the evaluated detection techniques, only \texttt{Semantic Entropy} maintains a degree of relative stability, exhibiting more modest performance variations between the two evaluation frameworks.

\subsection{Impact of Few-Shot Examples on Evaluation Reliability}
Our analysis of few-shot versus zero-shot settings reveals three key patterns in how examples affect evaluation stability (Table \ref{tab:ablation_metrics_few_vs_zero_shot}).

\paragraph{Improved Metric Stability}
Few-shot settings consistently yield more reliable evaluations across metrics. For \textsc{Llama}, the discrepancy between ROUGE and LLM-as-Judge narrows significantly with few-shot examples. For instance, \texttt{eRank}'s performance drop (for \textsc{Llama}) reduces from $-16.7\%$ in zero-shot to just $-4.2\%$ in few-shot settings. This suggests that few-shot examples help standardize response formats with more consistent evaluation.

\begin{table}[h]
    \centering
    \caption{\textbf{Few-shot examples reduce but don't eliminate evaluation biases.} Performance comparison showing relative differences between ROUGE and LLM-as-Judge in both settings.}

    \adjustbox{max width=0.5\textwidth}{
    \begin{tabular}{llrrr@{\hskip 2.0em}rrr} 
\toprule
\multirow{2}{*}{Model} & \multirow{2}{*}{Metric} & \multicolumn{3}{c}{Few-Shot} & \multicolumn{3}{c}{Zero-Shot} \\
 & & ROUGE & LLM & $\Delta$(\%) & ROUGE & LLM & $\Delta$(\%) \\

\midrule \midrule
  \multirow{5}{*}{\textsc{Llama}} & \texttt{Perplexity} & 0.783 & 0.784 & 0.0 & 0.715 & 0.725 & 1.5 \\
 & \texttt{LN-Entropy} & 0.738 & 0.759 & 2.8 & 0.547 & 0.617 & 12.8 \\

 & \texttt{SE} & 0.742 & 0.773 & 4.2 & 0.751 & 0.760 & 1.1 \\
 & \texttt{Eigenscore} & 0.761 & 0.747 & -1.9 & 0.763 & 0.686 & -10.0 \\
 & \texttt{eRank} & 0.707 & 0.678 & -4.2 & 0.765 & 0.638 & -16.7 \\
\cline{1-8}
 \multirow{5}{*}{\textsc{Mistral}} & \texttt{Perplexity} & 0.806 & 0.645 & -20.0 & 0.747 & 0.579 & -22.4 \\
 & \texttt{LN-Entropy} & 0.754 & 0.659 & -12.5 & 0.679 & 0.633 & -6.8 \\
 & \texttt{SE} & 0.750 & 0.732 & -2.4 & 0.782 & 0.712 & -8.9 \\
 & \texttt{Eigenscore} & 0.760 & 0.694 & -8.7 & 0.817 & 0.656 & -19.7 \\
 & \texttt{eRank} & 0.829 & 0.697 & -15.9 & 0.773 & 0.612 & -20.8 \\
\bottomrule
\end{tabular}

    }
    \label{tab:ablation_metrics_few_vs_zero_shot}
\end{table}

\paragraph{Model-Specific Effects}
The impact of few-shot examples varies notably between models. \textsc{Mistral} shows pronounced degradation in zero-shot settings, with performance drops up to $45.9\%$ (\texttt{Perplexity}), while \textsc{Llama} maintains more consistent performance, with some metrics showing minimal degradation. This variation suggests that the architecture and pre-training may influence the effectiveness of few-shot calibration.

\paragraph{Metric Robustness}
Different metrics show varying levels of stability across settings. \texttt{Semantic Entropy} maintains the most consistent performance in both settings, while traditional metrics like \texttt{Perplexity} or \texttt{LN-Entropy} show higher sensitivity to setting changes.

\paragraph{Implications}
 While few-shot examples generally improve evaluation reliability, the degree of improvement varies significantly across models and metrics. This suggests that robust hallucination detection systems should be validated under both conditions to ensure consistent performance across deployment scenarios. Of particular note is that few-shot examples reduce evaluation discrepancies by providing answer formats that more closely align with gold-standard responses. This indicates that some of the apparent improvements in few-shot settings may come from better format matching rather than enhanced factual assessment.

\subsection{Evaluating beyond ROUGE}
While ROUGE remains a widely adopted metric, its limitations underscore broader concerns about the reliability of reference-based evaluation methods. To assess whether alternative metrics fare better, we extended our analysis to several others frequently used or proposed for text evaluation, including BERTScore \cite{bertscore}, BLEU \cite{papineni-etal-2002-bleu}, SummaC \cite{laban-etal-2022-summac}, and UniEval-fact \cite{zhong-etal-2022-unieval}.
We evaluated these metrics in both few-shot and zero-shot settings, benchmarking their outputs against our LLM-as-Judge labels, which show strong alignment with human judgments (see Table~\ref{tab:human_agreements}).

\begin{table}[h]
    \centering
    \caption{\textbf{All metrics show limited alignment with human-like judgment, underscoring their shortcomings in capturing factual correctness.}
    Agreement of different correctness metrics with LLM-as-Judge labels in zero-shot settings. The results averaged across three QA datasets: NQ-Open, SQuAD, and TriviaQA.}

    \adjustbox{max width=0.5\textwidth}{
    
\begin{tabular}{llccccc}
\toprule
\textbf{Model} & \textbf{Metric} & \textbf{PRAUC} & \textbf{AUROC} & \textbf{F1} & \textbf{Precision} & \textbf{Recall} \\
\midrule
\multirow{5}{*}{\textsc{Llama}}
& BERTScore & 0.735 & 0.769 & 0.723 & 0.609 & 0.934 \\
& BLEU      & 0.758 & 0.624 & 0.673 & 0.539 & 0.982 \\
& ROUGE     & 0.891 & 0.878 & 0.812 & 0.728 & 0.926 \\
& SummaC    & 0.826 & 0.782 & 0.725 & 0.616 & 0.944 \\
& UniEval   & 0.828 & 0.830 & 0.762 & 0.739 & 0.804 \\
\midrule
\multirow{5}{*}{\textsc{Mistral}}
& BERTScore & 0.736 & 0.730 & 0.725 & 0.586 & 0.990 \\
& BLEU      & 0.799 & 0.682 & 0.712 & 0.573 & 0.996 \\
& ROUGE     & 0.865 & 0.825 & 0.757 & 0.629 & 0.971 \\
& SummaC    & 0.836 & 0.778 & 0.758 & 0.648 & 0.950 \\
& UniEval   & 0.720 & 0.706 & 0.693 & 0.674 & 0.746 \\
\bottomrule
\end{tabular}

    }
    \label{tab:label_metric_vs_llm_zero_shot}
\end{table}

\paragraph{Performance of Alternative Metrics}
As shown in Table~\ref{tab:label_metric_vs_llm_zero_shot}, these alternative metrics also exhibit substantial shortcomings in reliably detecting hallucinations in QA tasks, particularly under zero-shot conditions. For example, BERTScore—despite leveraging contextual embeddings—often failed to outperform simpler lexical metrics in aligning with our LLM-as-Judge labels. BLEU and UniEval-fact similarly demonstrated limited effectiveness.

\paragraph{Implications}
These results suggest that the inadequacies of ROUGE are not isolated, but indicative of a broader challenge: current reference-based metrics struggle to capture factual consistency, often favoring surface-level similarity or structural features such as length. Even when employing few-shot prompting (see Table~\ref{tab:label_metric_vs_llm_few_shot} in the Appendix \ref{app:labeling_metrics}), which can help with answer formatting, these metrics remain fundamentally constrained in their ability to assess factual correctness.

\section{The Length Factor: A Hidden Signal in Hallucination Detection}
Our analysis reveals a surprising and significant finding: response length alone serves as a powerful signal for detecting hallucinations. This discovery challenges conventional wisdom about hallucination detection and raises fundamental questions about the complexity needed in detection methods. Our investigation demonstrates that: \textbf{(1)} Simple length statistics can serve as surprisingly effective hallucination detectors, often matching or exceeding more sophisticated methods; \textbf{(2)} The strong influence of length on current evaluation methods raises concerns about their ability to assess factual correctness independently of response verbosity; \textbf{(3)} This relationship may provide insights into the underlying mechanisms of how LLMs generate incorrect information.

\subsection{Length Patterns in Hallucinated Responses}
\begin{figure}[h!]
    \centering
    \includegraphics[width=\linewidth]{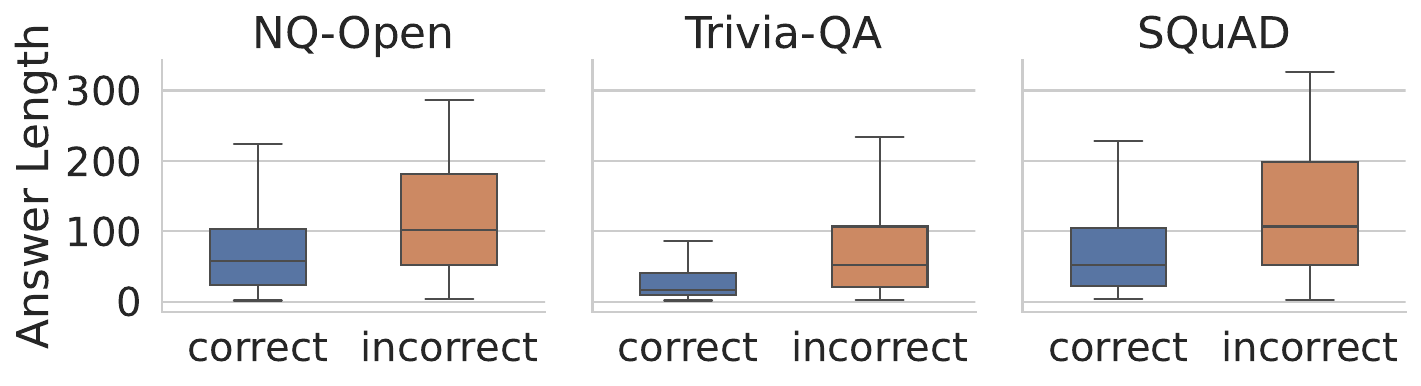}
    \caption{\textbf{Hallucinations have a distinct length signature in model outputs.} Distribution of answer lengths for \textsc{Mistral} in a few-shot settings with LLM-as-Judge labels, showing incorrect answers tend to be longer.}
    \label{fig:answer_length_distribution_few_shot}
\end{figure}

Analysis of response distributions using LLM-as-Judge labels reveals a striking pattern: hallucinated responses tend to be consistently longer and show greater length variance. This pattern holds true not only in our primary datasets (Figure~\ref{fig:answer_length_distribution_few_shot}) but also extends to the HaluEval dataset (Figure~\ref{fig:answer_length_distribution_halueval} in Appendix \ref{app:halueval_len_dist}), suggesting a fundamental relationship between verbosity and hallucination.

This tendency toward longer responses likely reflects two key mechanisms. First, models attempt to maintain coherence while generating incorrect information, leading to additional context and elaboration. Second, initial errors often cascade into further mistakes, creating a "snowball effect" of increasing verbosity \cite{zhang_how_2023}

\begin{figure}[h]
    \centering
    \includegraphics[width=0.9\linewidth]{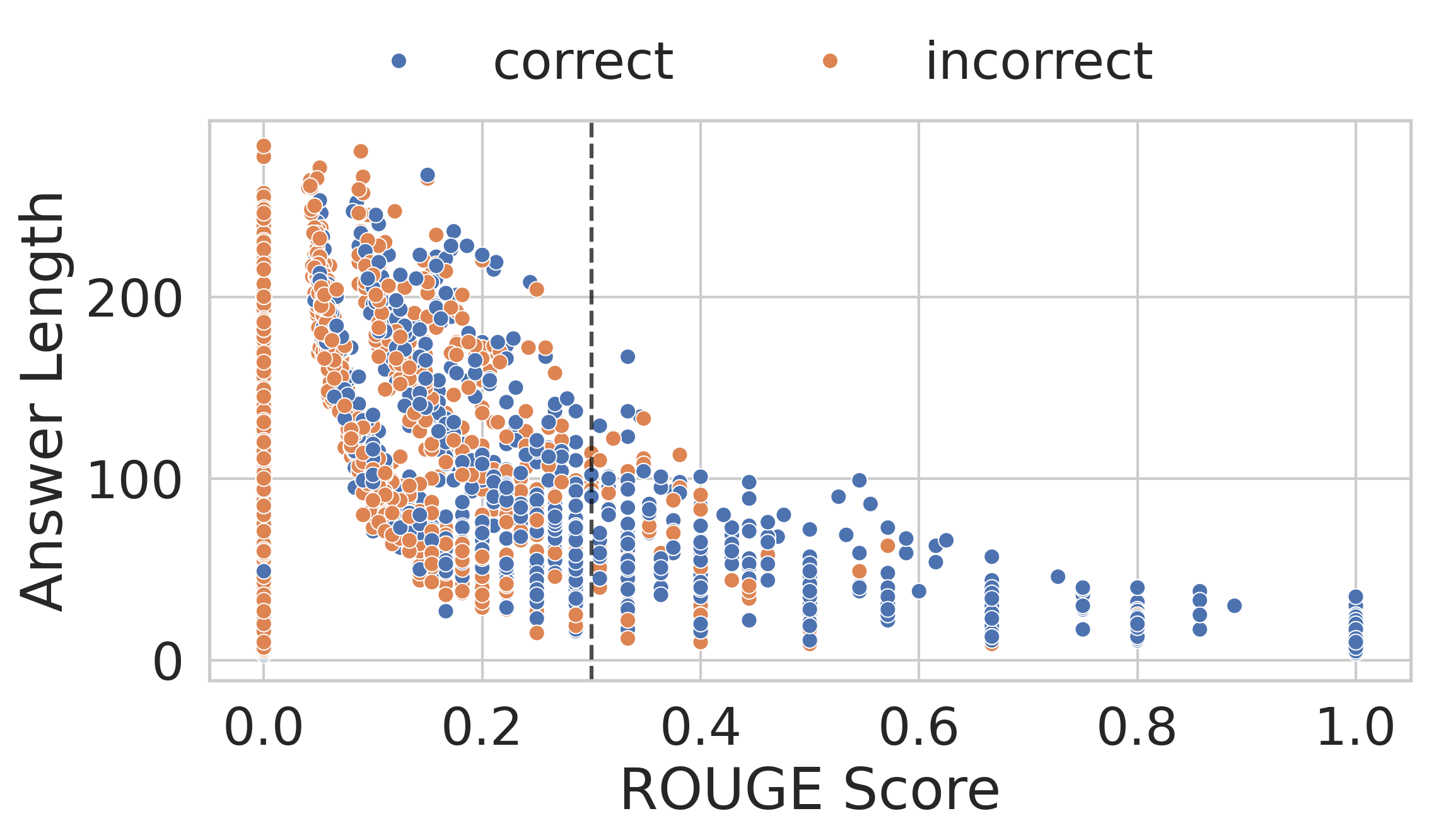}
    \caption{\textbf{ROUGE's bias against long responses undermines its reliability.} Distribution of answer length versus ROUGE score for \textsc{Mistral} in few-shot settings, revealing a strong correlation between length and ROUGE scores.}
    \label{fig:rouge_vs_length_mistral_few_shot}
\end{figure}

\subsection{Length Correlations with Existing Methods}
To quantify this relationship, we examined correlations between response length and various hallucination detection metrics. Our analysis reveals two critical findings. First, established methods show unexpectedly strong length correlations (see Table~\ref{tab:correlation_len}): \texttt{Eigenscore} and \texttt{eRank} exhibit particularly high correlations, suggesting these supposedly sophisticated methods may be primarily detecting length variations rather than semantic features. Second, ROUGE scores demonstrate systematic length bias: As shown in Figure~\ref{fig:rouge_vs_length_mistral_few_shot}, responses exceeding 100 tokens consistently receive scores below the 0.3 threshold, regardless of factual accuracy. This aligns with prior observations of hallucination snowballing \citep{zhang_how_2023}, where LLMs compound initial errors with additional mistakes.

\begin{table}[h]
    \centering
    \caption{\textbf{Sophisticated detection methods primarily capture length effects.} Pearson correlation coefficients between metrics and length, showing unexpectedly high values.}
    \adjustbox{max width=0.3\textwidth}{
    \begin{tabular}{lrr}
\toprule
Method & Llama & Mistral \\
\midrule
\texttt{LogDet} &	-0.185 &	0.311 \\
\texttt{Perplexity} &	0.841 &	-0.423 \\
\texttt{eRank} & 0.763 & 0.803 \\
\texttt{Eigenscore} & 0.826 & 0.894 \\
\texttt{LN-Entropy} & 0.305 & -0.753 \\
\texttt{Semantic Entropy} & 0.436 & 0.631 \\
\bottomrule
\end{tabular}

    }
    \label{tab:correlation_len}
\end{table}

These correlations raise fundamental questions about whether current hallucination detection methods are truly capturing semantic features or simply leveraging length-based patterns.

\subsection{Length as a Competitive Baseline}
Given these strong correlations, we developed three simple length-based metrics: the raw length of a single generation (\texttt{Len}), the average length across multiple generations (\texttt{Mean-Len}), and the standard deviation of lengths across generations (\texttt{Std-Len}).

Evaluation results (Table~\ref{tab:metric_results_len_baselines}) demonstrate that these straightforward metrics achieve surprisingly competitive performance. The \texttt{Mean-Len} metric matches or outperforms sophisticated approaches like \texttt{Eigenscore} and \texttt{LN-Entropy} across multiple datasets. Response length variability proves to be a key indicator, with \texttt{Std-Len} showing particular effectiveness at identifying hallucinations. Perhaps most surprisingly, even the simple \texttt{Len} metric achieves competitive performance, challenging the fundamental need for complex detection methods.

\begin{table}[h]
    \centering
    \caption{\textbf{Simple length-based metrics achieve competitive performance with sophisticated detection methods.} Hallucination detection performance (AUROC) compared across datasets and models using LLM-as-Judge since it shows better alignment with human judgements.}
    \adjustbox{max width=0.5\textwidth}{
    \begin{tabular}{llcccc}
\toprule
Model & Metric & \hspace{0.5em}NQ-Open\hspace{0.5em} & \hspace{0.5em}SQuAD\hspace{0.5em} & \hspace{0.5em}Trivia-QA\hspace{0.5em} & \hspace{0.5em}Mean\hspace{0.5em} \\  
\midrule \midrule
\multirow[t]{8}{*}{\textsc{Llama}} & \texttt{Perplexity} & 0.767 & 0.758 & 0.826 & 0.784 \\
 & \texttt{LN-Entropy} & 0.732 & 0.717 & 0.829 & 0.759 \\
 & \texttt{SE} & 0.730 & 0.741 & 0.849 & 0.773 \\
 & \texttt{Eigenscore} & 0.744 & 0.733 & 0.762 & 0.747 \\
 & \texttt{eRank} & 0.714 & 0.681 & 0.638 & 0.678 \\
 & \texttt{Len} & 0.686 & 0.687 & 0.640 & 0.671 \\
 & \texttt{Mean-Len} & 0.730 & 0.716 & 0.716 & 0.721 \\
 & \texttt{Std-Len} & 0.727 & 0.721 & 0.806 & 0.751 \\
\cline{1-6}
\multirow[t]{8}{*}{\textsc{Mistral}} & \texttt{Perplexity} & 0.632 & 0.636 & 0.637 & 0.635 \\
 & \texttt{LN-Entropy} & 0.619 & 0.667 & 0.692 & 0.659 \\
 & \texttt{SE} & 0.734 & 0.698 & 0.765 & 0.732 \\
 & \texttt{Eigenscore} & 0.686 & 0.691 & 0.706 & 0.694 \\
 & \texttt{eRank} & 0.698 & 0.690 & 0.703 & 0.697 \\
 & \texttt{Len} & 0.664 & 0.685 & 0.729 & 0.693 \\
 & \texttt{Mean-Len} & 0.683 & 0.705 & 0.750 & 0.713 \\
 & \texttt{Std-Len} & 0.577 & 0.589 & 0.665 & 0.610 \\
 \bottomrule
\end{tabular}

    }\label{tab:metric_results_len_baselines}
\end{table}

\subsection{The Repetition Experiment: Validating Length Effects}
To isolate the impact of length on evaluation metrics, we conducted a controlled experiment using systematic repetition. We modified model outputs by iteratively duplicating sentences while maintaining the same factual content. Results in Table~\ref{tab:dumb_mistral_few_shot} reveal a concerning trend: AUROC scores consistently improve with increased repetition, even though information content remains unchanged.
This experiment highlights a critical distinction: while verbose or repetitive responses may be inefficient, they aren't necessarily hallucinations if the core information is correct. However, current evaluation approaches, including both ROUGE and length-based metrics, fail to make this distinction.

\begin{table}[h]
\centering
\caption{\textbf{ROUGE scores can be manipulated through simple repetition.} AUROC measurements for \textsc{Mistral} when repeating the same content multiple times.}

\adjustbox{max width=0.5\textwidth}{
\begin{tabular}{lrrrr}
\toprule
Dataset & 0 & 1 & 2 & 4 \\
\midrule
NQ-Open & 0.852 & 0.935 (+9.7) & 0.955 (+12.1) & 0.964 (+13.1) \\
SQuAD & 0.842 & 0.894 (+6.2) & 0.909 (+8.0) & 0.948 (+12.6) \\
Trivia-QA & 0.843 & 0.901 (+6.9) & 0.907 (+7.6) & 0.919 (+9.0) \\
\bottomrule
\end{tabular}

}
\label{tab:dumb_mistral_few_shot}
\end{table}

\section{Discussion}
\label{sec:discussion} 
Our results reveal a clear misalignment between reference-based metrics, such as ROUGE, and human judgments in identifying hallucinations in QA.
Despite the short, focused nature of QA answers—where n‑gram overlap might seem sufficient—these metrics consistently reward fluent yet factually incorrect responses. While ROUGE is widely used, we further evaluated more sophisticated metrics — BERTScore, BLEU, and UniEval-fact — against judgments from a strong LLM-based evaluator, and similarly observed substantial disagreement, underscoring the limitations of these metrics in capturing factual consistency.
While careful prompt engineering or dataset-specific post-processing techniques might offer marginal improvements in ROUGE scores, these approaches often lack scalability and generalizability across different models and datasets. As demonstrated in our experiments, models frequently disregarded explicit brevity instructions (see prompts in Appendix \ref{app:prompts}), making the pursuit of an optimal, universally applicable prompt non-trivial endeavor.
The fundamental limitation of these reference-based metrics—their general insensitivity to factual veracity when masked by superficial lexical similarity—persists.
This is further underscored by our finding that simple response length can often be a more effective indicator of hallucinations than some sophisticated detection methods, questioning the current trajectory of detector development. These collective observations necessitate a shift towards more robust and semantically aware evaluation paradigms.

\section{Conclusions}
\label{sec:conclusions} %
We demonstrate that prevailing overlap‑based metrics systematically overestimate hallucination detection performance in QA, leading to illusory progress. LLM-as-Judge evaluation, validated against human judgments, exposes steep performance drops across all methods when judged by factual accuracy. Moreover, because simple signals like answer length can match complex detectors, we caution against over‑engineering: effective baselines are essential for meaningful advancement.

\section*{Limitations}
While our study provides valuable insights into the limitations of ROUGE for hallucination detection, several constraints should be acknowledged. First, our analysis primarily focuses on a subset of LLMs and datasets, which may not fully capture the diversity of models and tasks in the field. Consequently, the generalizability of our findings to other contexts remains to be validated. Second, although we propose response length as a simple yet effective heuristic for detecting hallucinations, this approach may not account for nuanced cases where longer responses are factually accurate. Additionally, our reliance on LLM-as-Judge as a benchmark for human-aligned evaluation, while more robust than ROUGE, is not without its own biases and limitations. Future work should explore alternative evaluation metrics and expand the scope of analysis to include a broader range of models and datasets. Finally, while our controlled experiments highlight the potential for manipulation of ROUGE scores, further research is needed to develop metrics that are both robust to such manipulations and aligned with human judgment. The primary risk is that over-reliance on length-based heuristics and potentially biased human-aligned metrics could lead to inaccurate assessments of hallucination detection methods, resulting in the deployment of LLMs that may not reliably ensure factual accuracy in high-stakes applications.

\bibliography{references-manual}

\begin{thebibliography}{44}
\expandafter\ifx\csname natexlab\endcsname\relax\def\natexlab#1{#1}\fi

\bibitem[{Arteaga et~al.(2024)Arteaga, Schön, and Pielawski}]{arteaga_hallucination_2024}
Gabriel~Y. Arteaga, Thomas~B. Schön, and Nicolas Pielawski. 2024.
\newblock \href {https://doi.org/10.48550/arXiv.2409.02976} {Hallucination {Detection} in {LLMs}: {Fast} and {Memory}-{Efficient} {Finetuned} {Models}}.
\newblock ArXiv:2409.02976 [cs].

\bibitem[{Azaria and Mitchell(2023)}]{azaria_internal_2023}
Amos Azaria and Tom Mitchell. 2023.
\newblock \href {http://arxiv.org/abs/2304.13734} {The {Internal} {State} of an {LLM} {Knows} {When} {It}'s {Lying}}.
\newblock ArXiv:2304.13734 [cs].

\bibitem[{Chen et~al.(2024)Chen, Liu, Chen, Gu, Wu, Tao, Fu, and Ye}]{chen_inside_2024}
Chao Chen, Kai Liu, Ze~Chen, Yi~Gu, Yue Wu, Mingyuan Tao, Zhihang Fu, and Jieping Ye. 2024.
\newblock \href {https://openreview.net/forum?id=Zj12nzlQbz} {{INSIDE}: {LLM}s' internal states retain the power of hallucination detection}.
\newblock In \emph{The Twelfth International Conference on Learning Representations}.

\bibitem[{Du et~al.(2024)Du, Xiao, and Li}]{du_haloscope_2024}
Xuefeng Du, Chaowei Xiao, and Sharon Li. 2024.
\newblock \href {https://proceedings.neurips.cc/paper_files/paper/2024/file/ba92705991cfbbcedc26e27e833ebbae-Paper-Conference.pdf} {Haloscope: Harnessing unlabeled llm generations for hallucination detection}.
\newblock In \emph{Advances in Neural Information Processing Systems}, volume~37, pages 102948--102972. Curran Associates, Inc.

\bibitem[{Duan et~al.(2024)Duan, Cheng, Wang, Zavalny, Wang, Xu, Kailkhura, and Xu}]{duan-etal-2024-shifting}
Jinhao Duan, Hao Cheng, Shiqi Wang, Alex Zavalny, Chenan Wang, Renjing Xu, Bhavya Kailkhura, and Kaidi Xu. 2024.
\newblock \href {https://doi.org/10.18653/v1/2024.acl-long.276} {Shifting attention to relevance: Towards the predictive uncertainty quantification of free-form large language models}.
\newblock In \emph{Proceedings of the 62nd Annual Meeting of the Association for Computational Linguistics (Volume 1: Long Papers)}, pages 5050--5063, Bangkok, Thailand. Association for Computational Linguistics.

\bibitem[{Dziri et~al.(2022)Dziri, Kamalloo, Milton, Zaiane, Yu, Ponti, and Reddy}]{dziri-etal-2022-faithdial}
Nouha Dziri, Ehsan Kamalloo, Sivan Milton, Osmar Zaiane, Mo~Yu, Edoardo~M. Ponti, and Siva Reddy. 2022.
\newblock \href {https://doi.org/10.1162/tacl_a_00529} {{F}aith{D}ial: A faithful benchmark for information-seeking dialogue}.
\newblock \emph{Transactions of the Association for Computational Linguistics}, 10:1473--1490.

\bibitem[{et~al.(2024)}]{openai_gpt-4_2024}
OpenAI et~al. 2024.
\newblock \href {https://doi.org/10.48550/arXiv.2303.08774} {{GPT}-4 {Technical} {Report}}.
\newblock ArXiv:2303.08774 [cs].

\bibitem[{Fadeeva et~al.(2023)Fadeeva, Vashurin, Tsvigun, Vazhentsev, Petrakov, Fedyanin, Vasilev, Goncharova, Panchenko, Panov, Baldwin, and Shelmanov}]{fadeeva-etal-2023-lm}
Ekaterina Fadeeva, Roman Vashurin, Akim Tsvigun, Artem Vazhentsev, Sergey Petrakov, Kirill Fedyanin, Daniil Vasilev, Elizaveta Goncharova, Alexander Panchenko, Maxim Panov, Timothy Baldwin, and Artem Shelmanov. 2023.
\newblock \href {https://doi.org/10.18653/v1/2023.emnlp-demo.41} {{LM}-polygraph: Uncertainty estimation for language models}.
\newblock In \emph{Proceedings of the 2023 Conference on Empirical Methods in Natural Language Processing: System Demonstrations}, pages 446--461, Singapore. Association for Computational Linguistics.

\bibitem[{Farquhar et~al.(2024)Farquhar, Kossen, Kuhn, and Gal}]{farquhar_detecting_2024}
Sebastian Farquhar, Jannik Kossen, Lorenz Kuhn, and Yarin Gal. 2024.
\newblock \href {https://doi.org/10.1038/s41586-024-07421-0} {Detecting hallucinations in large language models using semantic entropy}.
\newblock \emph{Nature}, 630(8017):625--630.
\newblock Publisher: Nature Publishing Group.

\bibitem[{Garrido et~al.(2023)Garrido, Balestriero, Najman, and Lecun}]{garrido_rankme_2023}
Quentin Garrido, Randall Balestriero, Laurent Najman, and Yann Lecun. 2023.
\newblock \href {https://doi.org/10.48550/arXiv.2210.02885} {{RankMe}: {Assessing} the downstream performance of pretrained self-supervised representations by their rank}.
\newblock ArXiv:2210.02885 [cs].

\bibitem[{Grattafiori(2024)}]{grattafiori_llama_2024}
Aaron et~al. Grattafiori. 2024.
\newblock \href {https://doi.org/10.48550/arXiv.2407.21783} {The {Llama} 3 {Herd} of {Models}}.
\newblock ArXiv:2407.21783 [cs].

\bibitem[{Honovich et~al.(2022)Honovich, Aharoni, Herzig, Taitelbaum, Kukliansy, Cohen, Scialom, Szpektor, Hassidim, and Matias}]{honovich_true_2022}
Or~Honovich, Roee Aharoni, Jonathan Herzig, Hagai Taitelbaum, Doron Kukliansy, Vered Cohen, Thomas Scialom, Idan Szpektor, Avinatan Hassidim, and Yossi Matias. 2022.
\newblock \href {https://doi.org/10.48550/arXiv.2204.04991} {{TRUE}: {Re}-evaluating {Factual} {Consistency} {Evaluation}}.
\newblock ArXiv:2204.04991 [cs].

\bibitem[{Huang et~al.(2025)Huang, Yu, Ma, Zhong, Feng, Wang, Chen, Peng, Feng, Qin, and Liu}]{lei_2025_survey}
Lei Huang, Weijiang Yu, Weitao Ma, Weihong Zhong, Zhangyin Feng, Haotian Wang, Qianglong Chen, Weihua Peng, Xiaocheng Feng, Bing Qin, and Ting Liu. 2025.
\newblock \href {https://doi.org/10.1145/3703155} {A survey on hallucination in large language models: Principles, taxonomy, challenges, and open questions}.
\newblock \emph{ACM Trans. Inf. Syst.}, 43(2).

\bibitem[{Jiang et~al.(2023)Jiang, Sablayrolles, Mensch, Bamford, Chaplot, Casas, Bressand, Lengyel, Lample, Saulnier, Lavaud, Lachaux, Stock, Scao, Lavril, Wang, Lacroix, and Sayed}]{jiang_mistral_2023}
Albert~Q. Jiang, Alexandre Sablayrolles, Arthur Mensch, Chris Bamford, Devendra~Singh Chaplot, Diego de~las Casas, Florian Bressand, Gianna Lengyel, Guillaume Lample, Lucile Saulnier, Lélio~Renard Lavaud, Marie-Anne Lachaux, Pierre Stock, Teven~Le Scao, Thibaut Lavril, Thomas Wang, Timothée Lacroix, and William~El Sayed. 2023.
\newblock \href {https://doi.org/10.48550/arXiv.2310.06825} {Mistral {7B}}.
\newblock ArXiv:2310.06825 [cs].

\bibitem[{Joshi et~al.(2017)Joshi, Choi, Weld, and Zettlemoyer}]{joshi_triviaqa_2017}
Mandar Joshi, Eunsol Choi, Daniel~S. Weld, and Luke Zettlemoyer. 2017.
\newblock \href {https://doi.org/10.48550/arXiv.1705.03551} {{TriviaQA}: {A} {Large} {Scale} {Distantly} {Supervised} {Challenge} {Dataset} for {Reading} {Comprehension}}.
\newblock ArXiv:1705.03551 [cs].

\bibitem[{Kang et~al.(2024)Kang, Blevins, and Zettlemoyer}]{kang2024}
Haoqiang Kang, Terra Blevins, and Luke Zettlemoyer. 2024.
\newblock \href {http://arxiv.org/abs/2402.10496} {Comparing hallucination detection metrics for multilingual generation}.

\bibitem[{Kossen et~al.(2024)Kossen, Han, Razzak, Schut, Malik, and Gal}]{kossen_semantic_2024}
Jannik Kossen, Jiatong Han, Muhammed Razzak, Lisa Schut, Shreshth Malik, and Yarin Gal. 2024.
\newblock \href {https://doi.org/10.48550/arXiv.2406.15927} {Semantic {Entropy} {Probes}: {Robust} and {Cheap} {Hallucination} {Detection} in {LLMs}}.
\newblock ArXiv:2406.15927 [cs].

\bibitem[{Kwiatkowski et~al.(2019)Kwiatkowski, Palomaki, Redfield, Collins, Parikh, Alberti, Epstein, Polosukhin, Devlin, Lee, Toutanova, Jones, Kelcey, Chang, Dai, Uszkoreit, Le, and Petrov}]{kwiatkowski_natural_2019}
Tom Kwiatkowski, Jennimaria Palomaki, Olivia Redfield, Michael Collins, Ankur Parikh, Chris Alberti, Danielle Epstein, Illia Polosukhin, Jacob Devlin, Kenton Lee, Kristina Toutanova, Llion Jones, Matthew Kelcey, Ming-Wei Chang, Andrew~M. Dai, Jakob Uszkoreit, Quoc Le, and Slav Petrov. 2019.
\newblock \href {https://doi.org/10.1162/tacl_a_00276} {Natural {Questions}: {A} {Benchmark} for {Question} {Answering} {Research}}.
\newblock \emph{Transactions of the Association for Computational Linguistics}, 7:452--466.
\newblock Place: Cambridge, MA Publisher: MIT Press.

\bibitem[{Laban et~al.(2022)Laban, Schnabel, Bennett, and Hearst}]{laban-etal-2022-summac}
Philippe Laban, Tobias Schnabel, Paul~N. Bennett, and Marti~A. Hearst. 2022.
\newblock \href {https://doi.org/10.1162/tacl_a_00453} {{S}umma{C}: Re-visiting {NLI}-based models for inconsistency detection in summarization}.
\newblock \emph{Transactions of the Association for Computational Linguistics}, 10:163--177.

\bibitem[{Li et~al.(2023)Li, Cheng, Zhao, Nie, and Wen}]{li_halueval_2023}
Junyi Li, Xiaoxue Cheng, Wayne~Xin Zhao, Jian-Yun Nie, and Ji-Rong Wen. 2023.
\newblock \href {https://doi.org/10.48550/arXiv.2305.11747} {{HaluEval}: {A} {Large}-{Scale} {Hallucination} {Evaluation} {Benchmark} for {Large} {Language} {Models}}.
\newblock ArXiv:2305.11747 [cs].

\bibitem[{Lin(2004)}]{lin_rouge_2004}
Chin-Yew Lin. 2004.
\newblock \href {https://aclanthology.org/W04-1013/} {{ROUGE}: {A} {Package} for {Automatic} {Evaluation} of {Summaries}}.
\newblock In \emph{Text {Summarization} {Branches} {Out}}, pages 74--81, Barcelona, Spain. Association for Computational Linguistics.

\bibitem[{Lin et~al.(2023)Lin, Trivedi, and Sun}]{lin_generating_2023}
Zhen Lin, Shubhendu Trivedi, and Jimeng Sun. 2023.
\newblock \href {https://doi.org/10.48550/ARXIV.2305.19187} {Generating with {Confidence}: {Uncertainty} {Quantification} for {Black}-box {Large} {Language} {Models}}.
\newblock Publisher: arXiv Version Number: 3.

\bibitem[{Malinin and Gales(2021)}]{malinin2021uncertainty}
Andrey Malinin and Mark Gales. 2021.
\newblock \href {https://openreview.net/forum?id=jN5y-zb5Q7m} {Uncertainty estimation in autoregressive structured prediction}.
\newblock In \emph{International Conference on Learning Representations}.

\bibitem[{Nguyen et~al.(2025)Nguyen, He, Gandre, Pasupulety, Shivakumar, and Lerman}]{nguyen2025smoothinghallucinationsmitigatingllm}
Hieu Nguyen, Zihao He, Shoumik~Atul Gandre, Ujjwal Pasupulety, Sharanya~Kumari Shivakumar, and Kristina Lerman. 2025.
\newblock \href {http://arxiv.org/abs/2502.11306} {Smoothing out hallucinations: Mitigating llm hallucination with smoothed knowledge distillation}.

\bibitem[{Nikitin et~al.(2024)Nikitin, Kossen, Gal, and Marttinen}]{kernel_language_entropy}
Alexander Nikitin, Jannik Kossen, Yarin Gal, and Pekka Marttinen. 2024.
\newblock \href {https://proceedings.neurips.cc/paper_files/paper/2024/file/10c456d2160517581a234dfde15a7505-Paper-Conference.pdf} {Kernel language entropy: Fine-grained uncertainty quantification for llms from semantic similarities}.
\newblock In \emph{Advances in Neural Information Processing Systems}, volume~37, pages 8901--8929. Curran Associates, Inc.

\bibitem[{Orgad et~al.(2024)Orgad, Toker, Gekhman, Reichart, Szpektor, Kotek, and Belinkov}]{orgad_llms_2024}
Hadas Orgad, Michael Toker, Zorik Gekhman, Roi Reichart, Idan Szpektor, Hadas Kotek, and Yonatan Belinkov. 2024.
\newblock \href {https://doi.org/10.48550/arXiv.2410.02707} {{LLMs} {Know} {More} {Than} {They} {Show}: {On} the {Intrinsic} {Representation} of {LLM} {Hallucinations}}.
\newblock ArXiv:2410.02707.

\bibitem[{Orgad et~al.(2025)Orgad, Toker, Gekhman, Reichart, Szpektor, Kotek, and Belinkov}]{orgad2025llms}
Hadas Orgad, Michael Toker, Zorik Gekhman, Roi Reichart, Idan Szpektor, Hadas Kotek, and Yonatan Belinkov. 2025.
\newblock \href {https://openreview.net/forum?id=KRnsX5Em3W} {{LLM}s know more than they show: On the intrinsic representation of {LLM} hallucinations}.
\newblock In \emph{The Thirteenth International Conference on Learning Representations}.

\bibitem[{Papineni et~al.(2002)Papineni, Roukos, Ward, and Zhu}]{papineni-etal-2002-bleu}
Kishore Papineni, Salim Roukos, Todd Ward, and Wei-Jing Zhu. 2002.
\newblock \href {https://doi.org/10.3115/1073083.1073135} {{B}leu: a method for automatic evaluation of machine translation}.
\newblock In \emph{Proceedings of the 40th Annual Meeting of the Association for Computational Linguistics}, pages 311--318, Philadelphia, Pennsylvania, USA. Association for Computational Linguistics.

\bibitem[{Qiu and Miikkulainen(2024)}]{semantic_density}
Xin Qiu and Risto Miikkulainen. 2024.
\newblock Semantic density: Uncertainty quantification for large language models through confidence measurement in semantic space.
\newblock In \emph{Advances in Neural Information Processing Systems}, volume~37, pages 134507--134533. Curran Associates, Inc.

\bibitem[{Rajpurkar et~al.(2018)Rajpurkar, Jia, and Liang}]{rajpurkar_know_2018}
Pranav Rajpurkar, Robin Jia, and Percy Liang. 2018.
\newblock \href {https://doi.org/10.48550/arXiv.1806.03822} {Know {What} {You} {Don}'t {Know}: {Unanswerable} {Questions} for {SQuAD}}.
\newblock ArXiv:1806.03822 [cs].

\bibitem[{Ren et~al.(2023)Ren, Luo, Zhao, Krishna, Saleh, Lakshminarayanan, and Liu}]{ren2023outofdistribution}
Jie Ren, Jiaming Luo, Yao Zhao, Kundan Krishna, Mohammad Saleh, Balaji Lakshminarayanan, and Peter~J Liu. 2023.
\newblock \href {https://openreview.net/forum?id=kJUS5nD0vPB} {Out-of-distribution detection and selective generation for conditional language models}.
\newblock In \emph{The Eleventh International Conference on Learning Representations}.

\bibitem[{Roy and Vetterli(2007)}]{roy_effective_2007}
Olivier Roy and Martin Vetterli. 2007.
\newblock The {Effective} {Rank}: a {Measure} of {Effective} {Dimensionality}.

\bibitem[{Sriramanan et~al.(2024{\natexlab{a}})Sriramanan, Bharti, Sadasivan, Saha, Kattakinda, and Feizi}]{llm_check}
Gaurang Sriramanan, Siddhant Bharti, Vinu~Sankar Sadasivan, Shoumik Saha, Priyatham Kattakinda, and Soheil Feizi. 2024{\natexlab{a}}.
\newblock \href {https://proceedings.neurips.cc/paper_files/paper/2024/file/3c1e1fdf305195cd620c118aaa9717ad-Paper-Conference.pdf} {Llm-check: Investigating detection of hallucinations in large language models}.
\newblock In \emph{Advances in Neural Information Processing Systems}, volume~37, pages 34188--34216. Curran Associates, Inc.

\bibitem[{Sriramanan et~al.(2024{\natexlab{b}})Sriramanan, Bharti, Sadasivan, Saha, Kattakinda, and Feizi}]{sriramanan_llm-check_2024}
Gaurang Sriramanan, Siddhant Bharti, Vinu~Sankar Sadasivan, Shoumik Saha, Priyatham Kattakinda, and Soheil Feizi. 2024{\natexlab{b}}.
\newblock \href {https://openreview.net/forum?id=LYx4w3CAgy&referrer=%5Bthe%20profile%20of%20Shoumik%20Saha%5D(%2Fprofile%3Fid%3D~Shoumik_Saha1)} {{LLM}-{Check}: {Investigating} {Detection} of {Hallucinations} in {Large} {Language} {Models}}.

\bibitem[{Su et~al.(2024)Su, Wang, Ai, HU, Wu, Zhou, and Liu}]{su_unsupervised_2024}
Weihang Su, Changyue Wang, Qingyao Ai, Yiran HU, Zhijing Wu, Yujia Zhou, and Yiqun Liu. 2024.
\newblock \href {https://doi.org/10.48550/arXiv.2403.06448} {Unsupervised {Real}-{Time} {Hallucination} {Detection} based on the {Internal} {States} of {Large} {Language} {Models}}.
\newblock ArXiv:2403.06448 [cs].

\bibitem[{Thakur et~al.(2025)Thakur, Choudhary, Ramayapally, Vaidyanathan, and Hupkes}]{thakur2025judgingjudgesevaluatingalignment}
Aman~Singh Thakur, Kartik Choudhary, Venkat~Srinik Ramayapally, Sankaran Vaidyanathan, and Dieuwke Hupkes. 2025.
\newblock \href {http://arxiv.org/abs/2406.12624} {Judging the judges: Evaluating alignment and vulnerabilities in llms-as-judges}.

\bibitem[{Wolf et~al.(2020)Wolf, Debut, Sanh, Chaumond, Delangue, Moi, Cistac, Rault, Louf, Funtowicz, Davison, Shleifer, von Platen, Ma, Jernite, Plu, Xu, Le~Scao, Gugger, Drame, Lhoest, and Rush}]{wolf_transformers_2020}
Thomas Wolf, Lysandre Debut, Victor Sanh, Julien Chaumond, Clement Delangue, Anthony Moi, Pierric Cistac, Tim Rault, Remi Louf, Morgan Funtowicz, Joe Davison, Sam Shleifer, Patrick von Platen, Clara Ma, Yacine Jernite, Julien Plu, Canwen Xu, Teven Le~Scao, Sylvain Gugger, Mariama Drame, Quentin Lhoest, and Alexander Rush. 2020.
\newblock \href {https://doi.org/10.18653/v1/2020.emnlp-demos.6} {Transformers: {State}-of-the-{Art} {Natural} {Language} {Processing}}.
\newblock In \emph{Proceedings of the 2020 {Conference} on {Empirical} {Methods} in {Natural} {Language} {Processing}: {System} {Demonstrations}}, pages 38--45, Online. Association for Computational Linguistics.

\bibitem[{Xu et~al.(2024)Xu, Jain, and Kankanhalli}]{xu_hallucination_2024}
Ziwei Xu, Sanjay Jain, and Mohan Kankanhalli. 2024.
\newblock \href {https://doi.org/10.48550/arXiv.2401.11817} {Hallucination is {Inevitable}: {An} {Innate} {Limitation} of {Large} {Language} {Models}}.
\newblock ArXiv:2401.11817.

\bibitem[{Zha et~al.(2023)Zha, Yang, Li, and Hu}]{zha-etal-2023-alignscore}
Yuheng Zha, Yichi Yang, Ruichen Li, and Zhiting Hu. 2023.
\newblock \href {https://doi.org/10.18653/v1/2023.acl-long.634} {{A}lign{S}core: Evaluating factual consistency with a unified alignment function}.
\newblock In \emph{Proceedings of the 61st Annual Meeting of the Association for Computational Linguistics (Volume 1: Long Papers)}, pages 11328--11348, Toronto, Canada. Association for Computational Linguistics.

\bibitem[{Zhang et~al.(2023)Zhang, Press, Merrill, Liu, and Smith}]{zhang_how_2023}
Muru Zhang, Ofir Press, William Merrill, Alisa Liu, and Noah~A. Smith. 2023.
\newblock \href {https://doi.org/10.48550/arXiv.2305.13534} {How {Language} {Model} {Hallucinations} {Can} {Snowball}}.
\newblock ArXiv:2305.13534 [cs].

\bibitem[{Zhang et~al.(2020)Zhang, Kishore, Wu, Weinberger, and Artzi}]{bertscore}
Tianyi Zhang, Varsha Kishore, Felix Wu, Kilian~Q. Weinberger, and Yoav Artzi. 2020.
\newblock Bertscore: Evaluating text generation with {BERT}.
\newblock In \emph{8th International Conference on Learning Representations, {ICLR} 2020, Addis Ababa, Ethiopia, April 26-30, 2020}.

\bibitem[{Zheng et~al.(2023{\natexlab{a}})Zheng, Chiang, Sheng, Zhuang, Wu, Zhuang, Lin, Li, Li, Xing, Zhang, Gonzalez, and Stoica}]{zheng_judging_2023}
Lianmin Zheng, Wei-Lin Chiang, Ying Sheng, Siyuan Zhuang, Zhanghao Wu, Yonghao Zhuang, Zi~Lin, Zhuohan Li, Dacheng Li, E.~Xing, Haotong Zhang, Joseph~E. Gonzalez, and Ion Stoica. 2023{\natexlab{a}}.
\newblock \href {https://www.semanticscholar.org/paper/Judging-LLM-as-a-judge-with-MT-Bench-and-Chatbot-Zheng-Chiang/a0a79dad89857a96f8f71b14238e5237cbfc4787} {Judging {LLM}-as-a-judge with {MT}-{Bench} and {Chatbot} {Arena}}.
\newblock \emph{ArXiv}.

\bibitem[{Zheng et~al.(2023{\natexlab{b}})Zheng, Chiang, Sheng, Zhuang, Wu, Zhuang, Lin, Li, Li, Xing et~al.}]{zheng2023judging}
Lianmin Zheng, Wei-Lin Chiang, Ying Sheng, Siyuan Zhuang, Zhanghao Wu, Yonghao Zhuang, Zi~Lin, Zhuohan Li, Dacheng Li, Eric Xing, et~al. 2023{\natexlab{b}}.
\newblock Judging llm-as-a-judge with mt-bench and chatbot arena.
\newblock \emph{Advances in Neural Information Processing Systems}, 36:46595--46623.

\bibitem[{Zhong et~al.(2022)Zhong, Liu, Yin, Mao, Jiao, Liu, Zhu, Ji, and Han}]{zhong-etal-2022-unieval}
Ming Zhong, Yang Liu, Da~Yin, Yuning Mao, Yizhu Jiao, Pengfei Liu, Chenguang Zhu, Heng Ji, and Jiawei Han. 2022.
\newblock \href {https://doi.org/10.18653/v1/2022.emnlp-main.131} {Towards a unified multi-dimensional evaluator for text generation}.
\newblock In \emph{Proceedings of the 2022 Conference on Empirical Methods in Natural Language Processing}, pages 2023--2038, Abu Dhabi, United Arab Emirates. Association for Computational Linguistics.

\end{thebibliography}
\bibliographystyle{acl_natbib}

\appendix

\section*{Appendix}
\addcontentsline{toc}{section}{Appendix}

\section{Licenses and Computational Resources}
\label{app:licenses_resources}

\subsection{Datasets, models license}
The datasets and models used in this study are subject to specific licenses. NQ-Open, TriviaQA, and SQuAD are available under licenses that permit academic use. The \textsc{Llama3.1-8B-Instruct} and \textsc{Mistral-7B-Instruct-v0.3} models are open-source and can be accessed under their respective licenses, which allow for research and non-commercial use.\footnote{For detailed license information, please refer to the respective dataset and model documentation.}

\subsection{Hardware Specifications}
We generated data using Nvidia A40 with 40GB VRAM. For the remaining computations, we used CPU.

\section{Human Involvement and Ethics}
\label{app:ethics}
\subsection{Annotator Recruitment and Consent}
Participants were recruited through personal networks (friends and acquaintances) and participated voluntarily without financial compensation. They were informed of the study’s purpose and data usage beforehand. Verbal consent was obtained, and no personally identifiable information was collected. Participants had the right to withdraw at any time.

\subsection{Demographics}
All annotators were residents of Poland. No systematic collection of age, gender, or other demographic information was conducted.

\section{Use of AI Assistance}
AI assistants such as ChatGPT were utilized in various aspects of the research, including coding, data analysis, and writing tasks. These tools helped to automate repetitive tasks, generate initial drafts, and assist in exploring potential solutions. However, all AI-generated outputs were reviewed and refined by researchers to ensure accuracy and coherence.

\section{Prompts}
\label{app:prompts}
We used the following prompt formats to elicit responses from the models:

\begin{itemize}
    \item \textbf{QA (Zero-shot)}: Minimal prompt with no examples (Listing~\ref{lst:zeroshot})
    \item \textbf{QA (Few-shot)}: Adapted from \cite{kossen_semantic_2024}, includes multiple QA examples (Listing~\ref{lst:qa})
    \item \textbf{LLM-as-Judge}: Evaluation prompt with correctness labels, adapted from \cite{orgad_llms_2024} (Listing~\ref{lst:judge})
\end{itemize}

\begin{lstlisting}[
    label=lst:zeroshot,
    caption={Zero-shot prompt template},
    frame=single,
    backgroundcolor=\color{gray!10},
    basicstyle=\ttfamily\small,
    breaklines=true
]
Answer the following question as briefly as possible.

Question: {question}
Answer:
\end{lstlisting}

\begin{lstlisting}[
    label=lst:qa,
    caption={QA (Few-shot) prompt template},
    frame=single,
    backgroundcolor=\color{gray!10},
    basicstyle=\ttfamily\small,
    breaklines=true
]
Answer the following question as briefly as possible.

Here are several examples:

Question: What is the capital of France?
Answer: Paris

Question: Who wrote Romeo and Juliet?
Answer: William Shakespeare

Question: What is the boiling point of water in Celsius?
Answer: 100

Question: How many continents are there on Earth?
Answer: Seven

Question: What is the fastest land animal?
Answer: Cheetah

Question: {question}
Answer:
\end{lstlisting}

\begin{lstlisting}[
    label=lst:judge,
    caption={LLM-as-Judge prompt template},
    frame=single,
    backgroundcolor=\color{gray!10},
    basicstyle=\ttfamily\small,
    breaklines=true
]
Answer the following question as briefly as possible.

Here are several examples:

Question: who is the young guitarist who played with Buddy Guy?
Ground Truth: Quinn Sullivan, Eric Gales
Model Answer: Ronnie Earl
Correctness: incorrect

Question: What is the name of the actor who plays Iron Man in the Marvel movies?
Ground Truth: Robert Downey Jr.
Model Answer: Robert Downey Jr. played the role of Tony Stark/Iron Man in the Marvel Cinematic Universe films.
Correctness: correct

Question: What is the capital of France?
Ground Truth: Paris
Model Answer: I don't have enough information to answer this question.
Correctness: refuse

Question: Who was the first person to walk on the moon?
Ground Truth: Neil Armstrong
Model Answer: I apologize, but I cannot provide an answer without verifying the historical facts.
Correctness: refuse

Question: {question}
Ground Truth: {gold}
Model Answer: {prediction}
Correctness:
\end{lstlisting}

\section{Additional Analysis of Human Evaluation}
\label{app:human_eval_analysis}
For the human evaluation component of our study (Section \ref{sec:human_eval}), we intentionally curated a dataset of instances where ROUGE and our LLM-as-Judge metric provided conflicting assessments regarding the presence of hallucinations. This targeted selection strategy was employed to enable a focused examination of ROUGE's specific failure modes. By concentrating on these points of disagreement, we aimed to gain deeper insights into the scenarios where ROUGE's reliance on lexical overlap demonstrably misaligns with human judgments of factual accuracy and overall response quality.

\section{Evaluation Metrics and Hallucination Detection}
\label{app:metrics}
\subsection{eRank}
\label{app:erank}
eRank leverages eigenvalue-based entropy estimation in hidden states:

\begin{equation}
    \texttt{eRank} = \exp\left(- \sum_{k=1}^{m} p_k \log p_k \right)
\end{equation}
where \(p_k = \frac{\lambda_k}{\sum_{j=1}^{m} \lambda_j}\), and \(\lambda_k\) are the eigenvalues of the covariance matrix \(\Sigma = {Z}^T {Z}\) computed on the hidden states $Z$.

We use Effective Rank (eRank) as a proxy for how “spread out” or “diverse” the final-layer hidden representations are (however, we can also the the hidden representation from middle-layer). Intuitively, if the model’s representation space collapses to fewer dimensions (i.e., low eRank), it may indicate that the model is relying on less context or ignoring crucial input signals—often manifesting as hallucinations. Conversely, a higher eRank suggests a richer, more nuanced encoding of the input, which typically correlates with more grounded and accurate responses. This approach builds on prior work \cite{sriramanan_llm-check_2024} (\texttt{LogDet}), which computes the log-determinant of the covariance matrix.

While initial evaluations under ROUGE suggested some promise, we found that eRank did not consistently correlate with hallucination rates across all datasets and settings when assessed using human-aligned metrics.
This 'negative results' illustrate how ROUGE's limitations can mislead method development.

\section{Understanding ROUGE's Failure Modes}
\label{app:rouge_failure_modes_analysis}
Through detailed error analysis, we identify three critical limitations in ROUGE's evaluation approach: (1) sensitivity to response length, (2) inability to handle semantic equivalence, and (3) over-reliance on exact lexical matches. Our analysis reveals that these limitations lead to both false negatives—factually correct responses marked as incorrect—and false positives—incorrect responses receiving high scores. As shown in Figure \ref{fig:fn_fp_barplot}, these errors occur frequently across different datasets and models.

\subsection{Length-Based Penalties}

\fbox{
    \parbox{0.45\textwidth}{
    \textbf{Question:} When was \textit{Pride and Prejudice} written? \\
    \textbf{Prediction:} ``Pride and Prejudice was written by Jane Austen and published in 1813.'' \\
    \textbf{Gold Answer:} ``1813'
    }
}

ROUGE systematically penalizes factually correct but verbose answers. In this example, despite providing accurate information with helpful context, the response receives a low score purely due to length mismatch. As shown in Figure~\ref{fig:rouge_vs_length_mistral_few_shot}, this bias affects longer responses regardless of their factual accuracy, with responses exceeding $100$ tokens consistently scoring below our $0.3$ threshold.  Notably, this is \textbf{the most frequent type of error} ROUGE makes.

\subsection{Semantic Equivalence Failures}

\fbox{
    \parbox{0.45\textwidth}{
    \textbf{Question:} What is one element a topographic map shows? \\
    \textbf{Prediction:} ``Elevation'' \\
    \textbf{Gold Answer:} ``Relief''
    }
}

ROUGE fails to recognize semantic equivalence between different phrasings. Here, despite "elevation" and "relief" being contextually equivalent terms in topography, ROUGE assigns a lower score due to lexical mismatch. This limitation systematically undervalues responses that use valid alternative terminology.

\subsection{False Lexical Matches}

\fbox{
    \parbox{0.45\textwidth}{
    \textbf{Question:} ``How many episodes of \textit{Grey's Anatomy} season 14?'' \\
    \textbf{Prediction:} ``23 episodes.'' \\
    \textbf{Gold Answer:} ``24 episodes.''
    }
}

ROUGE can assign high scores to factually incorrect answers that share surface structure with the reference. Despite the critical numerical error, the response receives a relatively high score due to matching surrounding words. This creates a dangerous bias toward structurally similar but factually wrong answers.

\section{Quantitative Results}
\label{app:full_results}

\subsection{QA Accuracy Across Settings}
\label{app:qa_acc}

Table \ref{tab:appendix_accuracy} presents the accuracies on the QA datasets. These accuracies are computed by selecting the most likely answer at a low temperature setting and comparing it to labels derived from either ROUGE or LLM-as-Judge evaluations.

\begin{table}[h]
\centering
\caption{Accuracies of different models, datasets, and prompts for the QA task.}

\adjustbox{max width=0.5\textwidth}{
\begin{tabular}{lllrrr}
\toprule
& & & & \multicolumn{2}{c}{Accuracy}\\
\cmidrule(lr){5-6}
Dataset & Model & Prompt & \# Refused & ROUGE & LLM \\
\midrule
NQ-Open & Llama & Few-Shot & 692 & 28.1\% & 29.2\% \\
NQ-Open & Llama & Zero-Shot & 139 & 24.2\% & 43.2\% \\
NQ-Open & Mistral & Few-Shot & 117 & 20.9\% & 35.8\% \\
NQ-Open & Mistral & Zero-Shot & 72 & 7.8\% & 39.0\% \\ \midrule
SQuAD & Llama & Few-Shot & 924 & 22.0\% & 18.3\% \\
SQuAD & Llama & Zero-Shot & 447 & 20.2\% & 25.0\% \\
SQuAD & Mistral & Few-Shot & 230 & 16.0\% & 22.6\% \\
SQuAD & Mistral & Zero-Shot & 116 & 5.8\% & 25.3\% \\ \midrule
Trivia-QA & Llama & Few-Shot & 95 & 63.7\% & 69.4\% \\
Trivia-QA & Llama & Zero-Shot & 39 & 58.8\% & 71.1\% \\
Trivia-QA & Mistral & Few-Shot & 11 & 53.8\% & 69.7\% \\
Trivia-QA & Mistral & Zero-Shot & 2 & 29.0\% & 64.8\% \\
\bottomrule
\end{tabular}

}
\label{tab:appendix_accuracy}
\end{table}

\subsection{Metric Evaluation: AUROC}
\label{app:auroc}Tables~\ref{tab:appendix_metrics_results_zero_shot_full} and~\ref{tab:appendix_metrics_results_few_shot_full} present comprehensive results comparing LLM-based and ROUGE-based evaluation metrics across three datasets: NQ-Open, SQuAD, and Trivia-QA. We evaluate nine different metrics using \textbf{AUROC} evaluation metric for both Llama and Mistral models under zero-shot and few-shot settings.

\begin{table*}[h]
    \centering
    \adjustbox{valign=c, max width=\textwidth, scale=1}{\begin{tabular}{lc|rrr|rrr|rrr|rrr}
\toprule
\multirow{2}{*}{Model} & \multirow{2}{*}{Metric} & \multicolumn{3}{c|}{NQ-Open} & \multicolumn{3}{c|}{SQuAD} & \multicolumn{3}{c|}{Trivia-QA} & \multicolumn{3}{c}{Mean} \\
 & & ROUGE & LLM & $\Delta$\% & ROUGE & LLM & $\Delta$\% & ROUGE & LLM & $\Delta$\% & ROUGE & LLM & $\Delta$\% \\
\midrule \midrule
Llama & Perplexity & 0.709 & 0.700 & -1.2 & 0.703 & 0.687 & -2.4 & 0.733 & 0.789 & 7.2 & 0.715 & 0.725 & 1.2 \\
Llama & LN-Entropy & 0.521 & 0.605 & 13.9 & 0.558 & 0.611 & 8.7 & 0.563 & 0.636 & 11.5 & 0.547 & 0.617 & 11.4 \\
Llama & SE & 0.778 & 0.742 & -4.8 & 0.707 & 0.705 & -0.2 & 0.769 & 0.832 & 7.6 & 0.751 & 0.760 & 0.9 \\
Llama & Eigenscore & 0.816 & 0.686 & -19.0 & 0.720 & 0.638 & -12.7 & 0.752 & 0.734 & -2.5 & 0.763 & 0.686 & -11.4 \\
Llama & eRank & 0.825 & 0.632 & -30.6 & 0.754 & 0.621 & -21.4 & 0.717 & 0.660 & -8.6 & 0.765 & 0.638 & -20.2 \\
Llama & Len & 0.834 & 0.616 & -35.3 & 0.777 & 0.622 & -24.9 & 0.760 & 0.691 & -10.0 & 0.790 & 0.643 & -23.4 \\
Llama & LogDet & 0.511 & 0.515 & 0.7 & 0.521 & 0.536 & 2.7 & 0.604 & 0.509 & -18.6 & 0.545 & 0.520 & -5.1 \\
Llama & Mean-Len & 0.825 & 0.654 & -26.1 & 0.743 & 0.643 & -15.7 & 0.771 & 0.743 & -3.8 & 0.780 & 0.680 & -15.2 \\
Llama & Std-Len & 0.711 & 0.644 & -10.5 & 0.664 & 0.627 & -6.0 & 0.759 & 0.754 & -0.7 & 0.711 & 0.675 & -5.7 \\ \midrule
Mistral & Perplexity & 0.852 & 0.584 & -45.9 & 0.516 & 0.500 & -3.2 & 0.843 & 0.627 & -34.4 & 0.737 & 0.570 & -27.8 \\
Mistral & LN-Entropy & 0.718 & 0.645 & -11.3 & 0.734 & 0.657 & -11.7 & 0.586 & 0.596 & 1.8 & 0.679 & 0.633 & -7.1 \\
Mistral & SE & 0.836 & 0.729 & -14.7 & 0.784 & 0.701 & -11.9 & 0.726 & 0.707 & -2.6 & 0.782 & 0.712 & -9.7 \\
Mistral & Eigenscore & 0.873 & 0.669 & -30.4 & 0.803 & 0.648 & -24.0 & 0.775 & 0.652 & -18.9 & 0.817 & 0.656 & -24.4 \\
Mistral & eRank & 0.925 & 0.678 & -36.4 & 0.518 & 0.511 & -1.3 & 0.851 & 0.645 & -31.9 & 0.765 & 0.611 & -23.2 \\
Mistral & Len & 0.934 & 0.634 & -47.2 & 0.860 & 0.624 & -37.8 & 0.929 & 0.673 & -37.9 & 0.908 & 0.644 & -41.0 \\
Mistral & LogDet & 0.628 & 0.508 & -23.6 & 0.562 & 0.518 & -8.5 & 0.843 & 0.606 & -39.2 & 0.678 & 0.544 & -23.8 \\
Mistral & Mean-Len & 0.890 & 0.643 & -38.4 & 0.828 & 0.626 & -32.2 & 0.875 & 0.667 & -31.3 & 0.864 & 0.645 & -34.0 \\
Mistral & Std-Len & 0.516 & 0.512 & -0.7 & 0.540 & 0.505 & -6.9 & 0.613 & 0.572 & -7.2 & 0.556 & 0.530 & -4.9 \\
\bottomrule
\end{tabular}
}
    \caption{Full comparison of LLM-based and ROUGE-based evaluation metrics across different datasets (NQ-Open, SQuAD, and Trivia-QA) for Llama and Mistral models in \textbf{zero-shot} setting using \textbf{AUROC} evaluation metric. The $\Delta\%$ columns show the relative percentage difference between LLM and ROUGE scores. Mean columns present the averaged scores across all datasets.}
\label{tab:appendix_metrics_results_zero_shot_full}
\end{table*}

\begin{table*}[h]
    \centering
    \adjustbox{valign=c, max width=\textwidth, scale=1}{\begin{tabular}{lc|rrr|rrr|rrr|rrr}
\toprule
\multirow{2}{*}{Model} & \multirow{2}{*}{Metric} & \multicolumn{3}{c|}{NQ-Open} & \multicolumn{3}{c|}{SQuAD} & \multicolumn{3}{c|}{Trivia-QA} & \multicolumn{3}{c}{Mean} \\
 & & ROUGE & LLM & $\Delta$\% & ROUGE & LLM & $\Delta$\% & ROUGE & LLM & $\Delta$\% & ROUGE & LLM & $\Delta$\% \\
\midrule \midrule
Llama & Perplexity & 0.814 & 0.767 & -6.1 & 0.736 & 0.758 & 2.9 & 0.800 & 0.826 & 3.1 & 0.783 & 0.784 & -0.0 \\
Llama & LN-Entropy & 0.753 & 0.732 & -2.9 & 0.663 & 0.717 & 7.5 & 0.799 & 0.829 & 3.6 & 0.738 & 0.759 & 2.7 \\
Llama & SE & 0.738 & 0.730 & -1.1 & 0.688 & 0.741 & 7.1 & 0.800 & 0.849 & 5.7 & 0.742 & 0.773 & 3.9 \\
Llama & Eigenscore & 0.813 & 0.744 & -9.3 & 0.725 & 0.733 & 1.2 & 0.745 & 0.762 & 2.3 & 0.761 & 0.746 & -1.9 \\
Llama & eRank & 0.794 & 0.714 & -11.2 & 0.708 & 0.681 & -4.0 & 0.620 & 0.638 & 2.8 & 0.707 & 0.678 & -4.1 \\
Llama & Len & 0.761 & 0.686 & -10.9 & 0.694 & 0.687 & -1.0 & 0.620 & 0.640 & 3.1 & 0.692 & 0.671 & -2.9 \\
Llama & LogDet & 0.729 & 0.690 & -5.6 & 0.659 & 0.636 & -3.7 & 0.590 & 0.618 & 4.5 & 0.659 & 0.648 & -1.6 \\
Llama & Mean-Len & 0.799 & 0.730 & -9.4 & 0.713 & 0.716 & 0.4 & 0.681 & 0.716 & 4.8 & 0.731 & 0.721 & -1.4 \\
Llama & Std-Len & 0.777 & 0.727 & -7.0 & 0.705 & 0.721 & 2.2 & 0.783 & 0.806 & 2.9 & 0.755 & 0.751 & -0.6 \\ \midrule
Mistral & Perplexity & 0.804 & 0.632 & -27.1 & 0.782 & 0.636 & -23.0 & 0.744 & 0.637 & -16.7 & 0.777 & 0.635 & -22.3 \\
Mistral & LN-Entropy & 0.727 & 0.619 & -17.4 & 0.785 & 0.667 & -17.7 & 0.750 & 0.692 & -8.3 & 0.754 & 0.659 & -14.5 \\
Mistral & SE & 0.772 & 0.734 & -5.3 & 0.737 & 0.698 & -5.6 & 0.741 & 0.765 & 3.1 & 0.750 & 0.732 & -2.6 \\
Mistral & Eigenscore & 0.789 & 0.686 & -15.0 & 0.775 & 0.691 & -12.2 & 0.717 & 0.706 & -1.5 & 0.760 & 0.694 & -9.6 \\
Mistral & eRank & 0.874 & 0.698 & -25.1 & 0.829 & 0.690 & -20.1 & 0.786 & 0.703 & -11.8 & 0.830 & 0.697 & -19.0 \\
Mistral & Len & 0.879 & 0.664 & -32.2 & 0.857 & 0.685 & -25.1 & 0.858 & 0.729 & -17.7 & 0.865 & 0.693 & -25.0 \\
Mistral & LogDet & 0.737 & 0.663 & -11.2 & 0.687 & 0.631 & -8.9 & 0.612 & 0.630 & 2.9 & 0.679 & 0.641 & -5.7 \\
Mistral & Mean-Len & 0.834 & 0.683 & -22.1 & 0.822 & 0.705 & -16.5 & 0.806 & 0.750 & -7.4 & 0.821 & 0.713 & -15.3 \\
Mistral & Std-Len & 0.609 & 0.577 & -5.6 & 0.629 & 0.589 & -6.8 & 0.663 & 0.665 & 0.3 & 0.634 & 0.610 & -4.0 \\
\bottomrule
\end{tabular}
}
    \caption{Full comparison of LLM-based and ROUGE-based evaluation metrics across different datasets (NQ-Open, SQuAD, and Trivia-QA) for Llama and Mistral models in \textbf{few-shot} setting using \textbf{AUROC} evaluation metric. The $\Delta\%$ columns show the relative percentage difference between LLM and ROUGE scores. Mean columns present the averaged scores across all datasets.}
\label{tab:appendix_metrics_results_few_shot_full}
\end{table*}

\subsection{Metric Evaluation: PR-AUC}
\label{app:prauc}
Tables~\ref{tab:appendix_prauc_metrics_results_zero_shot_full} and~\ref{tab:appendix_prauc_metrics_results_few_shot_full} provide PR-AUC scores under the same conditions.

\begin{table*}[h]
    \centering
    \adjustbox{valign=c, max width=\textwidth, scale=1}{\begin{tabular}{lc|rrr|rrr|rrr|rrr}
\toprule
\multirow{2}{*}{Model} & \multirow{2}{*}{Metric} & \multicolumn{3}{c|}{NQ-Open} & \multicolumn{3}{c|}{SQuAD} & \multicolumn{3}{c|}{Trivia-QA} & \multicolumn{3}{c}{Mean} \\
 & & ROUGE & LLM & $\Delta$\% & ROUGE & LLM & $\Delta$\% & ROUGE & LLM & $\Delta$\% & ROUGE & LLM & $\Delta$\% \\
\midrule \midrule
Llama & Perplexity & 0.833 & 0.680 & -22.4 & 0.863 & 0.823 & -4.8 & 0.594 & 0.514 & -15.6 & 0.763 & 0.672 & -14.3 \\
Llama & LN-Entropy & 0.717 & 0.611 & -17.4 & 0.793 & 0.773 & -2.6 & 0.570 & 0.652 & 12.5 & 0.693 & 0.679 & -2.5 \\
Llama & SE & 0.845 & 0.695 & -21.5 & 0.864 & 0.829 & -4.1 & 0.575 & 0.533 & -7.9 & 0.761 & 0.686 & -11.2 \\
Llama & Eigenscore & 0.850 & 0.670 & -26.8 & 0.866 & 0.809 & -7.1 & 0.565 & 0.574 & 1.6 & 0.760 & 0.684 & -10.8 \\
Llama & eRank & 0.782 & 0.607 & -28.9 & 0.820 & 0.783 & -4.6 & 0.674 & 0.760 & 11.3 & 0.759 & 0.717 & -7.4 \\
Llama & Len & 0.865 & 0.681 & -27.2 & 0.885 & 0.820 & -8.0 & 0.605 & 0.548 & -10.4 & 0.785 & 0.683 & -15.2 \\
Llama & LogDet & 0.852 & 0.659 & -29.2 & 0.873 & 0.810 & -7.8 & 0.602 & 0.562 & -7.1 & 0.776 & 0.677 & -14.7 \\
Llama & Mean-Len & 0.851 & 0.658 & -29.3 & 0.870 & 0.808 & -7.7 & 0.573 & 0.568 & -0.9 & 0.765 & 0.678 & -12.6 \\
Llama & Std-Len & 0.825 & 0.647 & -27.6 & 0.846 & 0.802 & -5.5 & 0.562 & 0.570 & 1.4 & 0.744 & 0.673 & -10.6 \\ \midrule
Mistral & Perplexity & 0.664 & 0.536 & -23.8 & 0.951 & 0.754 & -26.0 & 0.690 & 0.752 & 8.3 & 0.768 & 0.681 & -13.8 \\
Mistral & LN-Entropy & 0.882 & 0.664 & -32.8 & 0.920 & 0.790 & -16.4 & 0.625 & 0.633 & 1.3 & 0.809 & 0.696 & -16.0 \\
Mistral & SE & 0.956 & 0.725 & -31.8 & 0.964 & 0.819 & -17.7 & 0.808 & 0.510 & -58.3 & 0.909 & 0.685 & -35.9 \\
Mistral & Eigenscore & 0.957 & 0.698 & -37.1 & 0.965 & 0.804 & -20.0 & 0.818 & 0.544 & -50.3 & 0.913 & 0.682 & -35.8 \\
Mistral & eRank & 0.658 & 0.506 & -30.0 & 0.955 & 0.755 & -26.4 & 0.534 & 0.704 & 24.2 & 0.716 & 0.655 & -10.7 \\
Mistral & Len & 0.964 & 0.682 & -41.4 & 0.973 & 0.803 & -21.0 & 0.849 & 0.536 & -58.4 & 0.929 & 0.674 & -40.3 \\
Mistral & LogDet & 0.964 & 0.699 & -37.9 & 0.950 & 0.753 & -26.1 & 0.847 & 0.550 & -54.1 & 0.920 & 0.667 & -39.4 \\
Mistral & Mean-Len & 0.958 & 0.671 & -42.8 & 0.966 & 0.786 & -22.9 & 0.833 & 0.548 & -52.1 & 0.919 & 0.668 & -39.3 \\
Mistral & Std-Len & 0.891 & 0.583 & -52.8 & 0.889 & 0.724 & -22.7 & 0.755 & 0.605 & -24.8 & 0.845 & 0.637 & -33.4 \\
\bottomrule
\end{tabular}
}
    \caption{Full comparison of LLM-based and ROUGE-based evaluation metrics across different datasets (NQ-Open, SQuAD, and Trivia-QA) for Llama and Mistral models in \textbf{zero-shot} setting using \textbf{PR-AUC} evaluation metric. The $\Delta\%$ columns show the relative percentage difference between LLM and ROUGE scores. Mean columns present the averaged scores across all datasets.}
\label{tab:appendix_prauc_metrics_results_zero_shot_full}
\end{table*}

\begin{table*}[h]
    \centering
    \adjustbox{valign=c, max width=\textwidth, scale=1}{\begin{tabular}{lc|rrr|rrr|rrr|rrr}
\toprule
\multirow{2}{*}{Model} & \multirow{2}{*}{Metric} & \multicolumn{3}{c|}{NQ-Open} & \multicolumn{3}{c|}{SQuAD} & \multicolumn{3}{c|}{Trivia-QA} & \multicolumn{3}{c}{Mean} \\
 & & ROUGE & LLM & $\Delta$\% & ROUGE & LLM & $\Delta$\% & ROUGE & LLM & $\Delta$\% & ROUGE & LLM & $\Delta$\% \\
\midrule \midrule
Llama & Perplexity & 0.844 & 0.824 & -2.4 & 0.861 & 0.891 & 3.4 & 0.551 & 0.502 & -9.8 & 0.752 & 0.739 & -2.9 \\
Llama & LN-Entropy & 0.810 & 0.796 & -1.8 & 0.828 & 0.874 & 5.3 & 0.525 & 0.522 & -0.5 & 0.721 & 0.731 & 1.0 \\
Llama & SE & 0.814 & 0.802 & -1.5 & 0.842 & 0.879 & 4.3 & 0.536 & 0.506 & -6.1 & 0.731 & 0.729 & -1.1 \\
Llama & Eigenscore & 0.829 & 0.802 & -3.4 & 0.852 & 0.876 & 2.7 & 0.511 & 0.542 & 5.7 & 0.731 & 0.740 & 1.7 \\
Llama & eRank & 0.746 & 0.726 & -2.8 & 0.711 & 0.762 & 6.8 & 0.679 & 0.737 & 7.9 & 0.712 & 0.742 & 4.0 \\
Llama & Len & 0.834 & 0.806 & -3.5 & 0.856 & 0.884 & 3.1 & 0.522 & 0.571 & 8.7 & 0.737 & 0.754 & 2.8 \\
Llama & LogDet & 0.817 & 0.800 & -2.1 & 0.859 & 0.882 & 2.6 & 0.526 & 0.582 & 9.6 & 0.734 & 0.755 & 3.4 \\
Llama & Mean-Len & 0.825 & 0.798 & -3.4 & 0.852 & 0.878 & 2.9 & 0.509 & 0.553 & 7.9 & 0.729 & 0.743 & 2.5 \\
Llama & Std-Len & 0.820 & 0.794 & -3.2 & 0.846 & 0.873 & 3.1 & 0.526 & 0.524 & -0.3 & 0.731 & 0.730 & -0.1 \\ \midrule
Mistral & Perplexity & 0.506 & 0.520 & 2.7 & 0.624 & 0.673 & 7.4 & 0.740 & 0.778 & 4.9 & 0.623 & 0.657 & 5.0 \\
Mistral & LN-Entropy & 0.508 & 0.505 & -0.6 & 0.587 & 0.615 & 4.5 & 0.759 & 0.825 & 8.0 & 0.618 & 0.648 & 4.0 \\
Mistral & SE & 0.872 & 0.754 & -15.7 & 0.898 & 0.843 & -6.5 & 0.609 & 0.538 & -13.3 & 0.793 & 0.712 & -11.8 \\
Mistral & Eigenscore & 0.873 & 0.738 & -18.4 & 0.902 & 0.842 & -7.2 & 0.598 & 0.567 & -5.5 & 0.791 & 0.716 & -10.4 \\
Mistral & eRank & 0.515 & 0.526 & 2.0 & 0.855 & 0.789 & -8.4 & 0.606 & 0.736 & 17.8 & 0.659 & 0.684 & 3.8 \\
Mistral & Len & 0.897 & 0.735 & -22.1 & 0.918 & 0.848 & -8.3 & 0.687 & 0.530 & -29.7 & 0.834 & 0.704 & -20.0 \\
Mistral & LogDet & 0.895 & 0.734 & -21.9 & 0.869 & 0.793 & -9.5 & 0.673 & 0.561 & -19.8 & 0.812 & 0.696 & -17.1 \\
Mistral & Mean-Len & 0.879 & 0.734 & -19.7 & 0.907 & 0.844 & -7.5 & 0.629 & 0.548 & -14.7 & 0.805 & 0.709 & -14.0 \\
Mistral & Std-Len & 0.827 & 0.683 & -20.9 & 0.873 & 0.808 & -8.0 & 0.546 & 0.608 & 10.1 & 0.749 & 0.700 & -6.3 \\
\bottomrule
\end{tabular}
}
    \caption{Full comparison of LLM-based and ROUGE-based evaluation metrics across different datasets (NQ-Open, SQuAD, and Trivia-QA) for Llama and Mistral models in \textbf{few-shot} setting using \textbf{PR-AUC} evaluation metric. The $\Delta\%$ columns show the relative percentage difference between LLM and ROUGE scores. Mean columns present the averaged scores across all datasets.}
\label{tab:appendix_prauc_metrics_results_few_shot_full}
\end{table*}

\section{Ground Truth Labeling Metrics}
\label{app:labeling_metrics}
To evaluate and compare automatic labeling strategies, we examined the agreement between various evaluation metrics and the LLM-as-Judge annotations (Table \ref{tab:label_metric_vs_llm_few_shot}). This analysis provides insight into the reliability of proxy labeling methods for hallucination detection.

\begin{table}[h!]
    \centering
    \caption{Few-Shot Evaluation Metrics agreement with LLM-as-Judge labels. The results averaged across three QA datasets: NQ-Open, SQuAD, and TriviaQA.}
    \adjustbox{max width=0.5\textwidth}{
    \begin{tabular}{lcccccc}
\toprule
\textbf{Model} & \textbf{Metric} & \textbf{PRAUC} & \textbf{AUROC} & \textbf{F1} & \textbf{Precision} & \textbf{Recall} \\
\midrule
\multirow{5}{*}{\textsc{Llama}} 
  & BERTScore & 0.810 & 0.848 & 0.776 & 0.742 & 0.859 \\
  & BLEU      & 0.775 & 0.536 & 0.699 & 0.576 & 0.976 \\
  & ROUGE     & 0.935 & 0.921 & 0.883 & 0.866 & 0.906 \\
  & SummaC    & 0.850 & 0.776 & 0.760 & 0.653 & 0.977 \\
  & UniEval   & 0.943 & 0.933 & 0.862 & 0.868 & 0.868 \\
\midrule
\multirow{5}{*}{\textsc{Mistral}} 
  & BERTScore & 0.764 & 0.770 & 0.749 & 0.637 & 0.958 \\
  & BLEU      & 0.784 & 0.627 & 0.707 & 0.581 & 0.987 \\
  & ROUGE     & 0.903 & 0.878 & 0.820 & 0.738 & 0.932 \\
  & SummaC    & 0.855 & 0.795 & 0.758 & 0.657 & 0.957 \\
  & UniEval   & 0.813 & 0.801 & 0.754 & 0.751 & 0.778 \\
\bottomrule
\end{tabular}

    }
    \label{tab:label_metric_vs_llm_few_shot}
\end{table}

\section{HaluEval Answer Length Distribution}
\label{app:halueval_len_dist}
\begin{figure}[h!]
    \centering
    \includegraphics[width=0.49\textwidth]{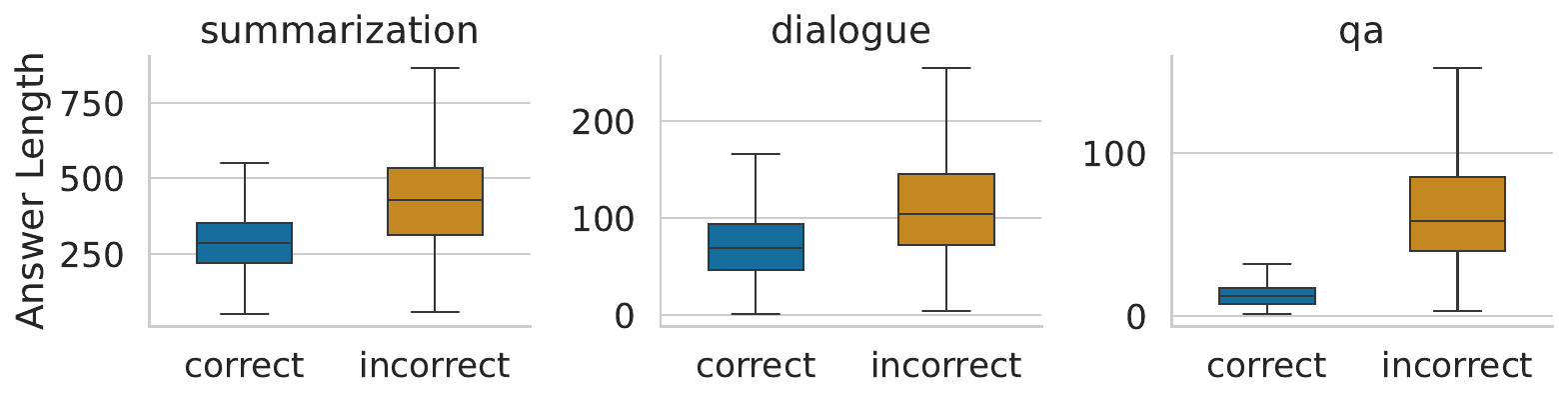}
    \caption{\textbf{Length-based hallucination patterns generalize across datasets.} Answer length distribution for HaluEval tasks, showing consistent patterns.}
    \label{fig:answer_length_distribution_halueval}
\end{figure}

Figure~\ref{fig:answer_length_distribution_halueval} illustrates answer lengths across the HaluEval dataset \cite{li_halueval_2023}.

\end{document}